\pdfoutput=1

\documentclass[11pt]{article}

\usepackage[preprint]{acl}

\usepackage{times}
\usepackage{latexsym}

\usepackage[T1]{fontenc}

\usepackage[utf8]{inputenc}

\usepackage{microtype}

\usepackage{inconsolata}

\usepackage{graphicx}
\usepackage{enumitem}
\usepackage{microtype}
\usepackage{subfigure}
\usepackage{amsmath}
\usepackage{booktabs} 
\usepackage{enumitem}
\usepackage{multirow} 
\usepackage{colortbl} 
\usepackage{float}
\usepackage{tabularx}
\usepackage{caption}
\usepackage{tcolorbox}
\usepackage{wrapfig}
\usepackage{pgfplots}
\usepackage{amsfonts}
\usepackage{booktabs}
\usepackage{subcaption}
\usepackage{amssymb}
\usepackage{titletoc}
\usepackage{tocbibind} 
\usepackage{hyperref}  
\usepackage{etoolbox}  
\hypersetup{
    colorlinks=true,       
    filecolor=magenta,     
}
\usepackage{graphicx}
\definecolor{myred}{RGB}{240, 183, 176}
\definecolor{myblue}{RGB}{184, 204, 226}
\title{Libra: \textbf{L}everaging Temporal \textbf{I}mages for \textbf{B}iomedical \textbf{R}adiology \textbf{A}nalysis}

\author{Xi Zhang\(^{1,2}\), Zaiqiao Meng\(^{1,2}\), Jake Lever\(^{1,2}\) \& Edmond S. L. Ho\\
  \(^{1}\)Information Retrieval Group, \(^{2}\)AI4BioMed Lab\\
  School of Computing Science\\
  University of Glasgow\\
  \texttt{\{X.Zhang.6\}@research.gla.ac.uk}\\
  \texttt{\{Zaiqiao.Meng,Jake.Lever,Shu-Lim.Ho\}@glasgow.ac.uk}
  \\
  \\
  \raisebox{-0.2\height}{\includegraphics[height=12pt]{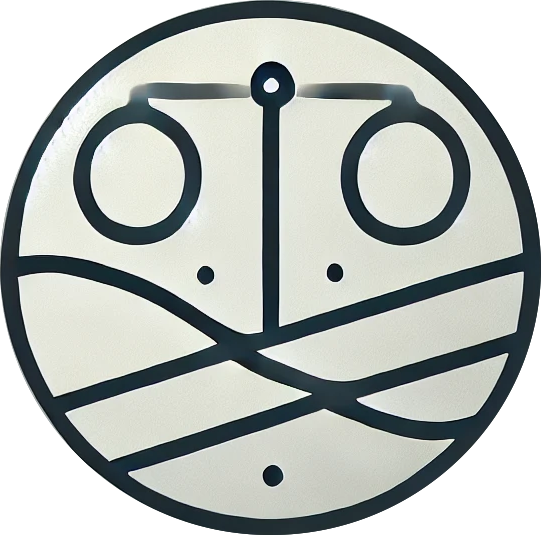}}
\href{https://x-izhang.github.io/Libra_v1.0}{\textcolor{blue}{https://x-izhang.github.io/Libra}}
}

\begin{document}
\maketitle
\begin{abstract}
Radiology report generation (RRG) requires advanced medical image analysis, effective temporal reasoning, and accurate text generation. While multimodal large language models (MLLMs) align with pre-trained vision encoders to enhance visual-language understanding, most existing methods rely on single-image analysis or rule-based heuristics to process multiple images, failing to fully leverage temporal information in multi-modal medical datasets. In this paper, we introduce \textbf{Libra}, a temporal-aware MLLM tailored for chest X-ray report generation. Libra combines a radiology-specific image encoder with a novel Temporal Alignment Connector (\textbf{TAC}), designed to accurately capture and integrate temporal differences between paired current and prior images. Extensive experiments on the MIMIC-CXR dataset demonstrate that Libra establishes a new state-of-the-art benchmark among similarly scaled MLLMs, setting new standards in both clinical relevance and lexical accuracy.
\end{abstract}

\section{Introduction}
Radiology reports are critical for biomedical radiology analysis, offering structured summaries of imaging studies such as chest X-rays (CXRs). Commonly divided into sections like \textit{Findings, Impression, Indication, Technique, Comparison}, and \textit{History} \citep{ganeshan2018structured}, these reports guide diagnostic and therapeutic decisions \citep{najjar2023redefining}. However, manually generating such reports is both complex and time-consuming. Automating radiology report generation (RRG) holds great promise for alleviating radiologist burnout, increasing efficiency, and improving communication \citep{zhang2020radiologyreportgenerationmeets}. Despite this, the intricate nature of medical imaging demands precise and detailed documentation, making RRG a challenging task.

\begin{figure}[t]
    \vspace{-5pt}
    \centering
    \includegraphics[width=\columnwidth]{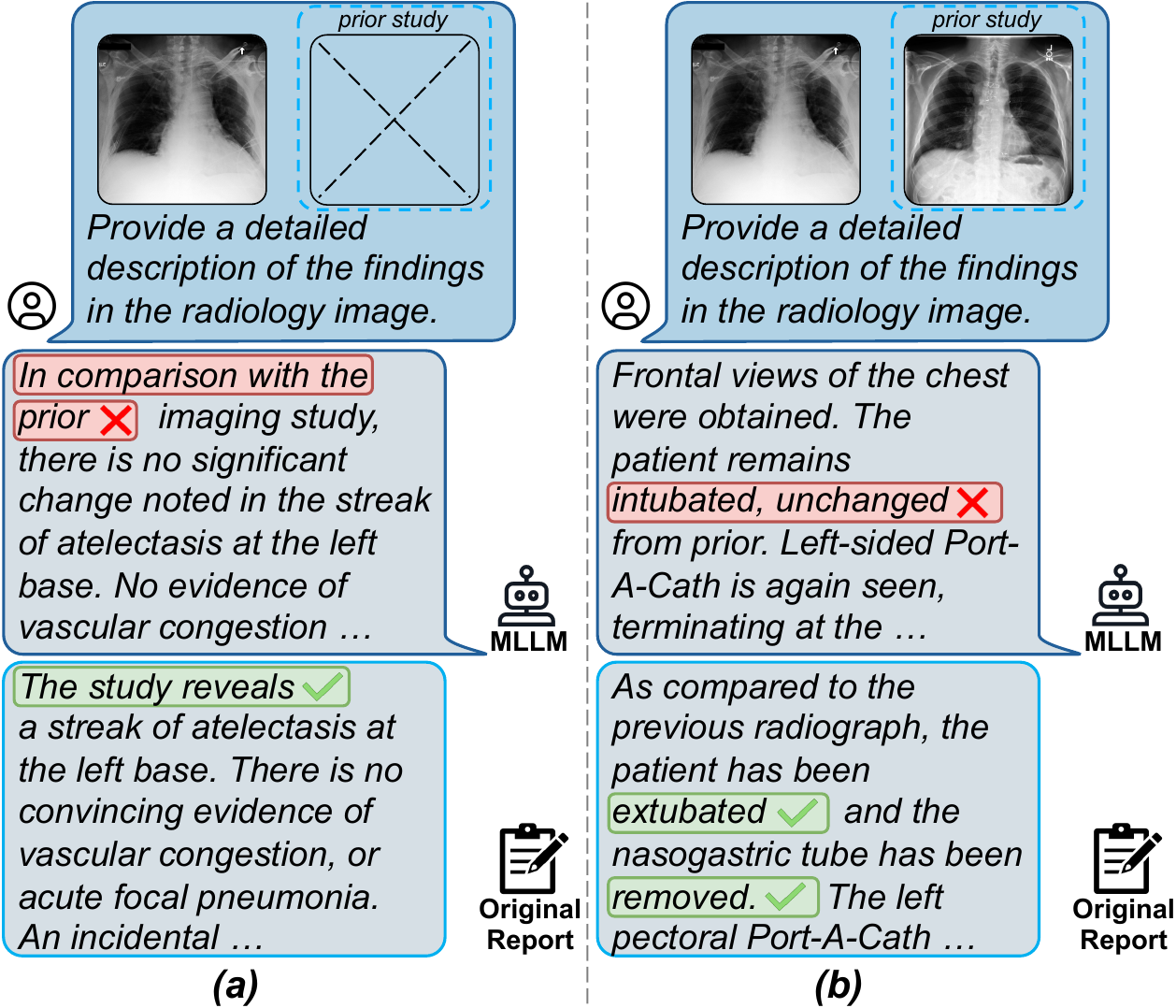}
    \vspace*{-20pt}
    \caption{Examples of hallucinations in RRG using the MLLM (MAIRA-1 \citep{hyland2024maira1specialisedlargemultimodal}). \textbf{\textit{(a)}} Single-image case: spurious references to nonexistent prior studies. \textbf{\textit{(b)}} Temporal image case: inaccurate interpretation of temporal changes when integrating prior studies.}
    \label{fig:case-problem}
    \vspace{-18pt}
\end{figure}

Recent advances in Multimodal Large Language Models (MLLMs), such as LLaVA \citep{liu2023visualinstructiontuning} and InstructBLIP \citep{dai2023instructblipgeneralpurposevisionlanguagemodels}, have demonstrated potential in vision-language tasks. However, their performance diminishes in specialised biomedical contexts due to the significant domain shift between general-purpose and medical image-text data \citep{tu2023generalistbiomedicalai}. These models often lack fine-grained detail for medical imaging tasks, resulting in surface-level understanding akin to layperson interpretations. While continued pre-training on medical datasets and domain-specific fine-tuning (e.g., LLaVA-Med \citep{li2023llavamedtraininglargelanguageandvision}, MedBLIP \citep{chen2023medblipbootstrappinglanguageimagepretraining}) improve performance, they still cannot fully capture the complexities of medical image analysis \citep{XIAO2025102888}.

One critical gap in the current MLLM-based approaches is their limited ability to incorporate temporal context, which is pivotal in clinical practice. Radiologists routinely compare current imaging results with prior studies to identify temporal changes, a process crucial for understanding disease progression and guiding treatment decisions. Indeed, the MIMIC-CXR database \citep{johnson2019mimic} reveals that \textbf{67\%} of patients underwent at least two studies at different time intervals, underscoring the necessity of temporal reasoning. However, most MLLMs designed for RRG tasks focus on single-image analysis, neglecting this temporal dimension \citep{zhang2024biomedclipmultimodalbiomedicalfoundation}. As illustrated in Figure~\ref{fig:case-problem}, MAIRA-1 \citep{hyland2024maira1specialisedlargemultimodal} introduces hallucinated prior references in single-image analysis and misinterprets temporal changes when integrating prior studies. 

Although recent models\footnote{Detailed related work is discussed in Appx.~\ref{related-work}, and our research objectives are explained in Appx.~\ref{research-object}.} like MedVersa \citep{zhou2024generalistlearnermultifacetedmedical} and MAIRA-2 \citep{bannur2024maira2groundedradiologyreport} have introduced multi-image processing, they do not explicitly model or extract temporal differences. Instead, they rely on inserting visual tokens from different studies at specific points within textual inputs and delegate the reasoning task to the LLM. Similarly, \citet{banerjee2024directpreferenceoptimizationsuppressing} and \citet{chaves2024clinicallyaccessibleradiologyfoundation} leverage GPT-4V \citep{openai2024gpt4technicalreport} to eliminate hallucinated references to prior studies in the dataset but lacks dedicated mechanisms for modelling temporal progression. Additionally, existing MLLMs often rely on embeddings from the last or penultimate layer of the image encoder \citep{Chen_2023,zhang2024visionlanguagemodelsvisiontasks}, primarily capturing global features. However, RRG tasks require fine-grained details\footnote{E.g., severity and temporal progression of findings.} \citep{Sloan_2024}, which a single-layer embedding often cannot fully represent \citep{jiang2024clipdinovisualencoders}. To tackle these limitations, we enhance MLLM temporal awareness for RRG tasks by addressing two main challenges:

\begin{itemize}[align=right,itemindent=2em,labelsep=2pt,labelwidth=1em,leftmargin=0pt,nosep,topsep=5pt]
    \setlength{\itemsep}{4pt}
    \item Designing robust MLLM architectures that seamlessly handle prior study references in RRG.
    \item The scarcity of effective feature alignment projectors in MLLMs capable of handling the high-granularity requirements of downstream tasks.
\end{itemize}

To overcome these gaps, we propose \textbf{Libra} (\textbf{L}everaging Temporal \textbf{I}mages for \textbf{B}iomedical \textbf{R}adiology \textbf{A}nalysis), a novel temporal-aware framework tailored for RRG tasks. Libra employs a pre-trained visual transformer encoder, RAD-DINO \citep{pérezgarcía2024raddinoexploringscalablemedical}, to generate robust image features, which are then refined using a new projector crafted for the temporal awareness, before being fed into the medical large language model (LLM), Meditron \citep{chen2023meditron70bscalingmedicalpretraining}. Through a two-stage training strategy, Libra aligns temporal visual features with the text embedding space, improving temporal coherence in RRG. 

Our modular approach integrates state-of-the-art open-source pre-trained models for medical image and text processing while introducing a dedicated temporal-aware adapter to align visual and textual modalities within the embedding space. This paper makes the following contributions:

\begin{itemize}[align=right,itemindent=2em,labelsep=2pt,labelwidth=1em,leftmargin=0pt,nosep,topsep=5pt,label=$\circ$]
    \setlength{\itemsep}{4pt}
    \item \textbf{Libra}, a temporal-aware MLLM designed to model temporal references and mitigate temporal hallucinations in RRG tasks.
    \item \textbf{Temporal Alignment Connector (TAC)}, comprising the Layerwise Feature Extractor (LFE) and Temporal Fusion Module (TFM), which extracts high-granularity image features from multiple encoder layers and integrates temporal references from the prior study when available.
    \item \textbf{Extensive evaluation} on the MIMIC-CXR dataset, achieving state-of-the-art results on average among similarly scaled MLLMs, with case analysis illustrating Libra's architectural benefits.
\end{itemize}

\section{Libra}\label{section:Libra}
\subsection{Model Architecture}
\begin{figure*}[t]
    \vspace{-10pt}
    \centering
    \includegraphics[width=0.78\textwidth,keepaspectratio]{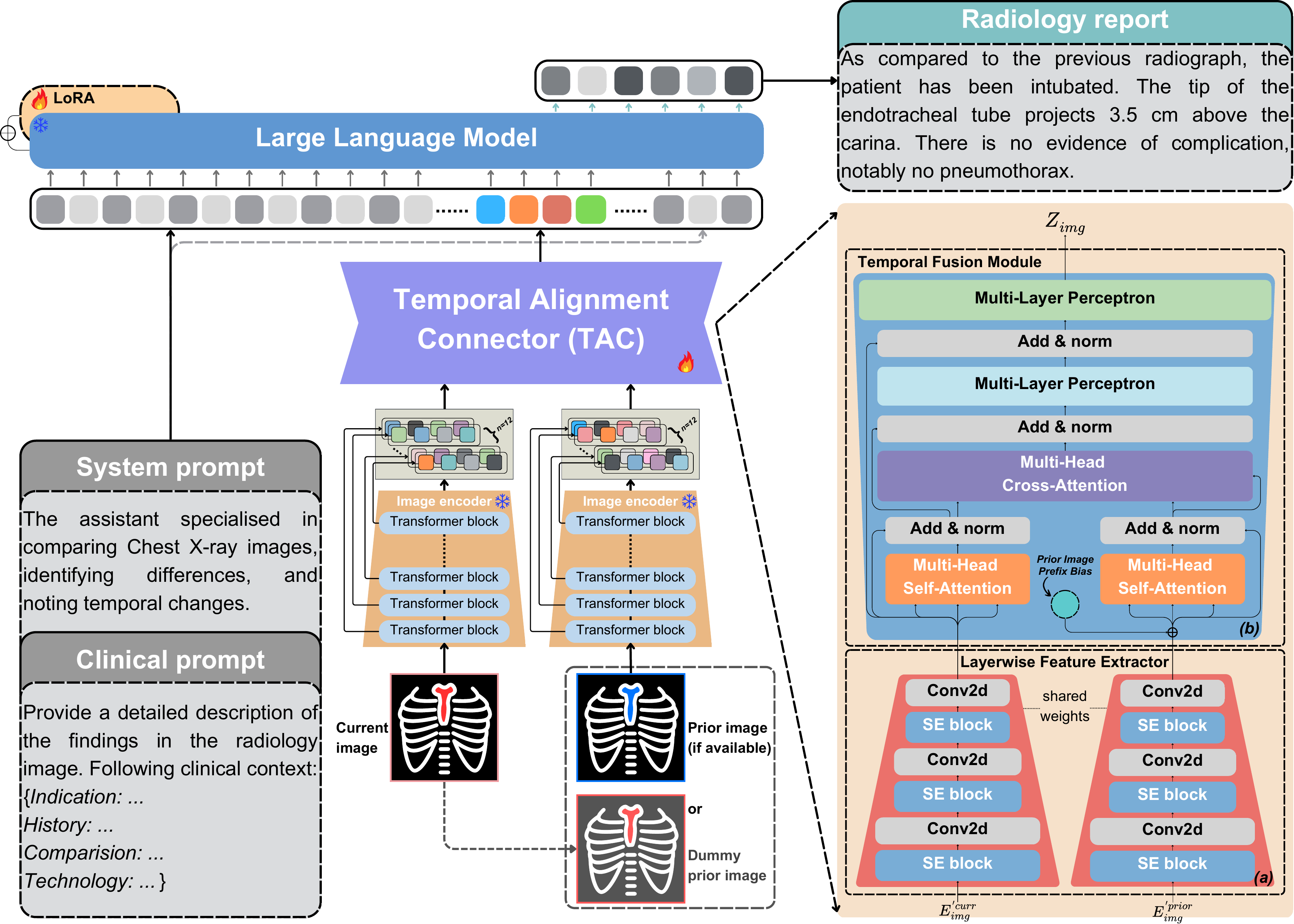}
    \vspace{-9pt}
    \caption{The overall architecture of Libra. The core component, the Temporal Alignment Connector (TAC), processes paired temporal images to enhance temporal reasoning. TAC consists of two key modules: \textbf{\textit{(a)}} the Layerwise Feature Extractor (LFE), which aggregates multi-layer image features from the image encoder, and \textbf{\textit{(b)}} the Temporal Fusion Module (TFM), which aligns the extracted features and integrates temporal differences before feeding them into the LLM. When no prior image is available, a dummy prior image is used to support temporal modelling, mitigate hallucinations, and prevent spurious references to nonexistent prior studies.}
    \label{fig:libra-framework}
    \vspace{-10pt}
\end{figure*}

Our Libra model follows the standard architecture of MLLMs, such as LLaVA \citep{liu2023visualinstructiontuning}, comprising an image encoder, a text decoder and a connector module to map visual features into the text embedding space. Figure \ref{fig:libra-framework} shows the overall architecture of Libra. Specifically, we utilise a frozen biomedical image encoder, i.e. RAD-DINO \citep{pérezgarcía2024raddinoexploringscalablemedical}, a visual transformer extensively pre-trained on medical scans using the DINOv2 image-only self-supervised learning approach \citep{oquab2024dinov2learningrobustvisual}. The text encoder is deployed by Meditron-7B \citep{chen2023meditron70bscalingmedicalpretraining}, which builds on Llama-2 and is further pre-trained on specialised medical corpora. 

To effectively connect the image encoder and LLM, we design a novel Temporal Alignment Connector (TAC) tailored to capture and integrate temporal information from paired images taken at different time points. Meanwhile, when no prior image is available, we employ a dummy prior image, which is simply a copy of the current image, to mitigate spurious references to nonexistent scans, as shown in Figure~\ref{fig:libra-framework} (bottom). This design enables Libra to effectively manage temporal data (e.g., stable, improved, worsening) and enhances its ability to generate accurate and coherent radiology reports.

\subsection{Temporal Alignment Connector}\label{Temporal Alignment Connector}
To address the challenges of integrating temporal information and aligning high-granularity visual features for RRG tasks, TAC bridges the image encoder and the LLM. It processes visual features from two temporal snapshots to produce a unified representation sensitive to temporal changes. As shown in Figure~\ref{fig:libra-framework} (right), TAC includes two key components: the \textit{Layerwise Feature Extractor}, which extracts high-granularity image representations, and the \textit{Temporal Fusion Module}, which integrates temporal references from the prior study.

\subsubsection{Layerwise Feature Extractor} 
To leverage abundant image feature representations encoded by a pre-trained image encoder, we extract image patch token features of all the hidden layers for a given pair of input images. By default, the RAD-DINO image encoder \citep{pérezgarcía2024raddinoexploringscalablemedical} has 12 hidden layers and processes 518 × 518 input images into 14 × 14 patches, generating 1,369 patch token sequences per hidden layer. Rather than relying on a single global feature token (e.g., the $[CLS]$ token), we collect same-dimensional patch embeddings from each layer per image, denoted as $E^{\text{img}}\in \mathbb{R}^{N \times D_{\text{img}}}$, where $N=1,369$ is the number of patch tokens and $D_{\text{img}}$ is the embedding dimension of the image encoder.

Then, these embeddings are concatenated across all layers as $E_{\text{img}}'=\{E_i^{\text{img}}\}_{i=1}^{n}$, where $n$ is the number of hidden layers.  Drawing from the progressive compression strategy in VGG \citep{simonyan2015deepconvolutionalnetworkslargescale}, our Layerwise Feature Extractor (LFE) reduces dimensionality across layers while preserving critical information. First, we utilise Squeeze-and-Excitation (SE) Networks \citep{hu2019squeezeandexcitationnetworks}, which construct informative features by integrating both spatial and channel-wise information within local receptive fields at each layer. The SE block is applied to obtain calibrated feature representations, using GELU~\citep{hendrycks2023gaussianerrorlinearunits} as the activation function. 

Next, we employ a specialised pointwise convolution module to align the feature spaces across different layers, using a depthwise 2D convolution with filters and stride of 1, without bias. The compressed features are represented as $A_{\text{img}} \sim Conv2d_j^{k}(SE_j^{k}(E_{\text{img}}'))$, where $k$ is the original layer number and $j$ is the layer number after compression. Following the size-reduction pattern of convolutional layers in VGG, the image features are compressed according to $\{k,j\} \in \{12,6,3,1\}$\footnote{Since RAD-DINO has $12$ hidden layers, the prime factorisation chain provides the factors as $\{12,6,3,1\}$.}. Through three stages of progressive compression, we obtain the final patch-level representation:

\vspace{-8pt}
\begingroup
\small
\begin{equation}
A_{\text{img}}'=Conv2d_6^{12}(SE_6^{12}(E_{\text{img}}'))
\end{equation}
\endgroup

\vspace{-10pt}
\begingroup
\small
\begin{equation}
A_{\text{img}}''=Conv2d_3^{6}(SE_3^{6}(A_{\text{img}}'))
\end{equation}
\endgroup

\vspace{-10pt}
\begingroup
\small
\begin{equation}
A_{\text{img}}=Conv2d_1^{3}(SE_1^{3}(A_{\text{img}}''))
\end{equation}
\endgroup

\vspace{-8pt}
For simplicity, we use ${LFE} (\cdot)$ to denote the above three stages of compression, which project a given input image $E_{\text{img}}'$ into its feature representation of the fixed dimension, $A_{\text{img}}\in \mathbb{R}^{ 1 \times N \times D_{\text{img}}}$:

\vspace{-8pt}
\begingroup
\small
\begin{equation}
A_{\text{img}}=LFE(E_{\text{img}}')
\end{equation}
\endgroup

\vspace{-8pt}
By progressively refining each image's representations through multiple stages,  the LFE generates a unified and compact feature set suitable for temporal alignment. This design ensures that both high-granularity and global context are retained, as illustrated in \textbf{\textit{(a)}} of Figure~\ref{fig:libra-framework}.

\subsubsection{Temporal Fusion Module} 

The Temporal Fusion Module (TFM) is inspired by the transformer decoder and is designed to integrate temporal information by leveraging prior images as auxiliary context. It takes as input a paired set of compressed features from both the current and prior images, denoted as $A_\text{img}^{\text{curr}}$ and $A_{\text{img}}^{\text{prior}}$, respectively, which are obtained after processing through the LFE. The temporal fusion process is defined as:

\vspace{-8pt}
\begingroup
\small
\begin{equation}
Z_{\text{img}}=TFM(A_\text{img}^{\text{curr}},A_{\text{img}}^{\text{prior}})
\end{equation}
\endgroup

\vspace{-8pt}
\noindent
where TFM learns to weigh the current image using prior image features, refining the representation to enhance temporal awareness. The resulting feature sequence, $Z_{\text{img}}\in \mathbb{R}^{N \times d}$, serves as the input to the LLM, where $N$ is the number of patch tokens and $d$ is the hidden dimension of the LLM. This process encapsulates the temporal evolution of the patient's condition, allowing the language model to generate accurate and contextually aware radiology reports.
\paragraph{Prior Image Prefix Bias} 
The dataset contains samples with and without a prior image. When a prior image is not available, we set $A_\text{img}^{\text{prior}} = A_\text{img}^{\text{curr}}$. However, this ``dummy prior image'' is indistinguishable from a true prior in raw features. To differentiate it, we add a trainable bias, as $b_\text{prior}$.

Following the attention scaling techniques for adjusting hidden space degrees of freedom with a chi-square distribution \citep{vaswani2017attention}, a nonlinear scaling function amplifies higher similarity values. The cosine similarity between the current and prior images is scaled with an exponent of $\sqrt[4]{d}$, where $d$ is the hidden dimension of the LLM:

\vspace{-8pt}
\begingroup
\small
\begin{equation}
b_\text{prior}'=b_\text{prior}\cdot\left(\frac{cos(A_\text{img}^{\text{curr}},A_\text{img}^{\text{prior}})+1}{2}\right)^{\sqrt[4]{d}}
\end{equation}
\endgroup

\vspace{-8pt}
\begingroup
\small
\begin{equation}
A_{\text{img}}'^{\text{prior}}=A_\text{img}^{\text{prior}}+ b_\text{prior}' 
\end{equation}
\endgroup

\vspace{-1.5mm}
This nonlinear scaling emphasises high similarity values, modulating the influence of prior image features. When no prior image is available, the high similarity score ensures that the effect of the dummy prior is adequately represented. This adjustment prevents samples with a dummy prior image from undergoing redundant rounds of parallel multi-head self-attention during subsequent propagation through the transformer blocks, in Figure~\ref{fig:libra-framework}.

\paragraph{Transformer Block} 
The Transformer Block in TFM follows the standard Transformer design but is optimized for handling temporal image pairs. It consists of multi-head self-attention ($SelfAttn$), multi-head cross-attention ($CrossAttn$), and two multi-layer perceptron ($MLP$) sub-layers. As illustrated in \textbf{\textit{(b)}} of Figure~\ref{fig:libra-framework}. The paired $(A_\text{img}^{\text{curr}},A_{\text{img}}'^{\text{prior}})$ are processed with layer normalization ($LN$) and residual connections:

\noindent
\vspace{-8pt}
\begingroup
\small
\begin{equation}
T_\text{curr}^{\text{self}}=LN(A_\text{img}^{\text{curr}}+ SelfAttn(A_\text{img}^{\text{curr}};A_\text{img}^{\text{curr}}))
\end{equation}
\endgroup

\noindent
\vspace{-12pt}
\begingroup
\small
\begin{equation}
T_\text{prior}^{\text{self}}=LN(A_{\text{img}}'^{\text{prior}}+ SelfAttn(A_{\text{img}}'^{\text{prior}};A_{\text{img}}'^{\text{prior}}))
\end{equation}
\endgroup

\noindent
\vspace{-12pt}
\begingroup
\small
\begin{equation}
T_\text{img}^{\text{cross}}=LN(T_\text{curr}^{\text{self}}+ CrossAttn(T_\text{curr}^{\text{self}};T_\text{prior}^{\text{self}}))
\end{equation}
\endgroup

\noindent
\vspace{-12pt}
\begingroup
\small
\begin{equation}
T_\text{img}^{\text{out}}=LN(A_\text{img}^{\text{curr}}+MLP_\text{attn}(T_\text{img}^{\text{cross}}))
\end{equation}
\endgroup

\noindent
\vspace{-12pt}
\begingroup
\small
\begin{equation}
Z_{\text{img}}=MLP_\text{final}(T_\text{img}^{\text{out}})
\end{equation}
\endgroup

\vspace{-8pt}
\noindent
where $MLP_\text{attn}$ is a simple neural network composed of two fully connected layers with GELU as the activation function. After that, the features are processed through $MLP_\text{final}$, a straightforward neural network consisting of four fully connected layers with the same activation function, but with hidden dimensions matching those of the LLM. 



\subsection{Prompt Design}
To enhance Libra's ability to perceive temporal changes and integrate medical information in RRG, we design a structured prompting strategy, consisting of a system prompt and a clinical prompt, as shown in Figure~\ref{fig:libra-framework} (left). The system prompt enables the LLM to recognise temporal variations,  while standard report sections (\textit{Indication}\, \textit{History}, \textit{Comparison}, \textit{Technique}) are integrated into the clinical prompt (see Appx.~\ref{prompt-example} for a detailed example). 

The full prompt is: ``Provide a detailed description of the findings in the radiology image. Following clinical context:$\{...\}$.'' There are datasets, e.g. MIMIC-CXR \citep{johnson2019mimic}, where the report sections are unavailable. For these datasets, we set the prompt as follows: ``Provide a detailed description of the findings in the radiology image.'' After tokenising and embedding prompts, the refined image features ($Z_{\text{img}}$) are inserted between the system prompt and clinical prompts.

\subsection{Temporal-aware Training}\label{section:temporal-training}
Libra focuses on frontal-view images, either posterior-anterior (PA) or anterior-posterior (AP), and targets the \textit{Findings} sections of RRG, as these contain the most direct clinical observations. It employs a two-stage training strategy, inspired by recent MLLM fine-tuning techniques \citep{mckinzie2024mm1methodsanalysis}, to progressively learn visual feature alignment and temporal information extraction. 

\vspace{-5pt}
\paragraph{Temporal Feature Alignment} In the first stage, the visual encoder and LLM weights are frozen, while the TAC is trained. This stage focuses on \textit{Findings} and \textit{Impression} generation from paired images and performing CXR-related visual question answering (VQA) tasks to extract high-quality image representations and capture temporal changes.

\vspace{-5pt}
\paragraph{Downstream Task Fine-tuning} In the second stage, we apply Low-Rank Adaptation (LoRA) \citep{hu2021loralowrankadaptationlarge} to fine-tune the LLM on the \textit{Findings} generation task, while keeping the visual encoder and TAC weights frozen. LoRA achieves performance comparable to full fine-tuning at a substantially lower computational cost. The detailed training configuration, including learning rate schedules and hyperparameters, is provided in Appx.~\ref{train-config}.

\begin{table*}[t]
    \vspace{-30pt}
    \begin{center}
    \scriptsize
    \setlength{\tabcolsep}{2pt} 
    \resizebox{0.85\linewidth}{!}{ 
    \begin{tabular}{l|cccccc|c}
        \toprule
        \textbf{Metric} & \textbf{LLaVA-Med$^\text{‡}$} & \textbf{CheXagent$^\text{‡}$}& \textbf{GPT-4V$^\text{‡}$} & \textbf{Med-PaLM} & \textbf{LLaVA-Rad} & \textbf{MAIRA-1} & \textbf{Libra} \scriptsize{(\%)}\\
        \midrule
        \textbf{Lexical:} & & & & & &  \\
        \hspace{1em} ROUGE-L & 27.6 & 21.5 & 13.2  & 27.5 & \underline{30.6} & 28.9  & \textbf{36.2} \scriptsize{(+18.3\%)} \\
        \hspace{1em} BLEU-1 & 35.4 & 16.9 & 16.4  & 32.3 & 38.1 & \underline{39.2} & \textbf{51.2} \scriptsize{(+30.6\%)} \\
        \hspace{1em} BLEU-4 & 14.9 & 4.7   & \underline{17.8}  & 11.5 & 15.4 & 14.2 & \textbf{24.3} \scriptsize{(+36.5\%)}\\
        \hspace{1em} METEOR & \underline{35.3} & -- & --  & -- & -- & 33.3   & \textbf{48.7}  \scriptsize{(+38.0\%)} \\ 
        \midrule
        \textbf{Clinical:} & & & & & &   \\
        \hspace{1em} RadGraph-F1 & 19.1 & -- & --  & \underline{26.7} & -- & 24.3 & \textbf{32.4} \scriptsize{(+21.3\%)}  \\
        \hspace{1em} RG$_{\text{ER}}$  & 23.8 & 20.5 & 13.2  & --   & 29.4 & \underline{29.6} & \textbf{36.9} \scriptsize{(+25.0\%)}  \\
        \hspace{1em} RadCliQ$_{\text{0}}$($\downarrow$)& 3.30 & -- & --    & -- & -- & \underline{3.10}  & \textbf{2.76} \scriptsize{(+11.0\%)} \\
        \hspace{1em} CheXbert vector & 36.9 & -- & --    & -- & -- & \underline{44.0}   & \textbf{46.3} \scriptsize{(+5.2\%)}  \\
        \hspace{1em} \textit{CheXpert-F1:} & & & & & & \\
        \hspace{2em} Micro-F1-14 & 42.7 & 39.3 & 35.5  & 53.6 & \textbf{57.3} & \underline{55.7}  & 55.3 \scriptsize{(-3.4\%)} \\
        \hspace{2em} Macro-F1-14 & 26.9 & 24.7 & 20.4 & \underline{39.8} & 39.5 & 38.6  & \textbf{40.2} \scriptsize{(+1.1\%)} \\
        \hspace{2em} Micro-F1-5  & 43.9 & 41.2 & 25.8  & \underline{57.9} & 57.4 & 56.0   & \textbf{58.9} \scriptsize{(+1.8\%)} \\
        \hspace{2em} Macro-F1-5 & 36.3 & 34.5 & 19.6  & \underline{51.6} & 47.7 & 47.7   & \textbf{52.6} \scriptsize{(+2.0\%)} \\  
        \bottomrule
    \end{tabular}
    }
    \end{center}
    \vspace{-13pt}
    \caption{\label{tab:libra-result}Findings generation performance on the MIMIC-CXR test split. $^\text{‡}$ denotes results from  \citet{chaves2024clinicallyaccessibleradiologyfoundation}, while `--' indicates missing data. The best performances in \textbf{bold}, and the second-best scores are \underline{underlined}. Metrics where lower values are better are marked with `$\downarrow$'. Percentage ($\%$) shows improvement over the best existing model.}
    \vspace{-13pt}
\end{table*}

\section{Experiments}
\vspace{-2pt}
\subsection{Task and Dataset}

\vspace{-1pt}
\paragraph{Task Description} We focus on generating the \textit{Findings} section of radiology reports for frontal CXRs, ensuring a fair comparison with prior work. The \textit{Findings} section provides radiologists' observations, encompassing both normal and abnormal findings. While additional sections like \textit{Indication} and \textit{Technique} primarily serve as routine records (e.g., clinical history or specific physician requests), they also assist the model in understanding temporal changes across images. Hence, we incorporate clinical instructions about the current image as prompts to guide Libra to complete the RRG task.

The most common CXR is frontal views, either PA or AP. Although lateral views are occasionally used to supplement anatomical assessments \citep{islam2023introductionmedicalimagingmodalities}, they are excluded in this study to maintain consistency with previous research on RRG tasks, such as \citet{chaves2024clinicallyaccessibleradiologyfoundation} and \citet{hyland2024maira1specialisedlargemultimodal}. Both current and prior images in our experiments exclusively utilise single frontal views.

\vspace{-5pt}
\paragraph{Dataset Description} Libra is trained and evaluated using the MIMIC-CXR dataset \citep{johnson2019mimic} and its derivative datasets, including Medical-Diff-VQA \citep{10.1145/3580305.3599819} and MIMIC-Ext-\textit{MIMIC-CXR-VQA} \citep{bae2023ehrxqamultimodalquestionanswering}. All datasets are split according to the official labels to prevent data leakage. Detailed dataset descriptions and preprocessing steps are in Appx.~\ref{detail-dataset}.

Following the dataset scaling law utilised in multi-stage MLLM fine-tuning methods \citep{zhu2023minigpt4enhancingvisionlanguageunderstanding}, we adopt a two-stage training strategy, as noted in Sec.~\ref{section:temporal-training}. The first stage trains TAC on $\sim$1.2M CXR-image text pairs from MIMIC-CXR and its derivatives, including \textit{Findings}, \textit{Impression}, and VQA tasks, enabling it to learn CXR token distributions and image-text relationships. The second stage fine-tunes the model on downstream tasks, refining the LLM to align high-granularity CXR features with the \textit{Findings} section of reports.
 
Beyond \textit{Findings} section generation, the first stage incorporates \textit{Impression} section and VQA tasks. The \textit{Impression} section, which summarises diagnoses and proposes further investigations \citep{babar2021evaluating}, facilitates alignment between CXRs and their textual descriptions. We use the same system and clinical prompts as for \textit{Findings}, replacing `Findings' with `Impression'. For VQA, the system prompts remain unchanged, while clinical prompts are adapted to address medical-specific questions, guiding caption generation. These VQA tasks refine the MLLM's biomedical vocabulary usage and strengthen image-text alignment.

\subsection{Evaluation Metrics}
We evaluate the generated reports using lexical and radiology-specific metrics, adhering to established protocols. Lexical metrics include ROUGE-L \citep{lin-2004-rouge}, BLEU-$\{1,4\}$ \citep{10.3115/1073083.1073135}, METEOR \citep{banerjee-lavie-2005-meteor} and BERT \citep{devlin2019bertpretrainingdeepbidirectional}. Radiology-specific metrics include RadGraph-F1 \citep{jain2021radgraphextractingclinicalentities}, RG$_{\text{ER}}$ \citep{delbrouck-etal-2022-improving}, F1-CheXpert \citep{irvin2019chexpertlargechestradiograph}, CheXbert vector similarity \citep{smit-etal-2020-combining}, and RadCliQ \citep{Yu2022.08.30.22279318} version $0$. 

These clinical metrics typically emphasise the accuracy of medical findings, prioritising the detection of clinically relevant entities. However, they do not evaluate the model's ability to capture temporal information. Therefore, we introduce the temporal entity F1 score ($F1_\text{temp}$) to assess this aspect.  In particular, the temporal entity F1 score specifically measures the accuracy of entities related to progression over time described in the report \footnote{Full metric descriptions, including $F1_\text{temp}$, are in Appx.~\ref{Metrics}.}.

\vspace{-5pt}
\paragraph{Temporal Entity F1} Building on the work of \citet{bannur2023learningexploittemporalstructure}, we set a reward list comprising common radiology-related keywords indicative of temporal changes. Temporal entities are then extracted from both the ground truth $(E_\text{gt})$ and the generated reports $(E_\text{gr})$ without applying stemming or lemmatization, preserving the precision of temporal descriptions. After extraction, we compute precision $(P_\text{temp})$ and recall $(R_\text{temp})$, which are subsequently used to calculate the $F1_\text{temp}$, defined as the harmonic mean of precision and recall \citep{pub.1012204240}, also known as the $F1$ score.

\vspace{-8pt}
\begingroup
\small
\begin{equation}
F1_\text{temp} = (1 + \beta^2) \cdot \frac{P_\text{temp} R_\text{temp}}{\beta^2 \cdot P_\text{temp} + R_\text{temp}}
\end{equation}
\endgroup

\vspace{-10pt}
\begingroup
\small
\begin{equation}
P_\text{temp} = \frac{\left|E_\text{gr} \cap E_\text{gt} \right|+ \epsilon}{\left| E_\text{gr} \right|+ \epsilon}
\end{equation}
\endgroup

\vspace{-10pt}
\begingroup
\small
\begin{equation}
R_\text{temp} = \frac{\left|E_\text{gr} \cap E_\text{gt} \right|+ \epsilon}{\left| E_\text{gt} \right|+ \epsilon}
\end{equation}
\endgroup

\vspace{-8pt}
\noindent
where $\epsilon$ is a small value, set to a default of $1 \times 10^{-10}$, to prevent division by zero (it is also added to the numerator for special cases where no temporal entities are present in the ground truth). 

\subsection{Baselines}
While the MIMIC-CXR dataset provides an ``official'' test split, strict comparisons with prior studies are challenging due to differences in inclusion criteria and pre-processing steps. For instance, \citet{Yu2022.08.30.22279318} and \citet{jeong2023multimodalimagetextmatchingimproves} included only one image per study, resulting in a test set of 1,597 samples, while \citet{Tanida_2023_CVPR} followed the Chest ImaGenome split \citep{wu2021chestimagenomedatasetclinical}. Such variations in test set distributions can significantly impact the reported results \citep{park2024m4cxrexploringmultitaskpotentials}. To ensure fairness~\footnote{The test set includes 2,461 frontal-view samples.}, we use a widely adopted test set focused on frontal-view CXRs, aligned with previous studies such as MAIRA-1 \citep{hyland2024maira1specialisedlargemultimodal} and LLaVA-Rad \citep{chaves2024clinicallyaccessibleradiologyfoundation}.
 
Recent concurrent work, such as M4CXR \citep{park2024m4cxrexploringmultitaskpotentials}, employs multi-turn chain-of-thought prompting \citep{wei2023chainofthoughtpromptingelicitsreasoning} for report generation, which differs from our task setup. Additionally, we do not compare with MAIRA-2 \citep{bannur2024maira2groundedradiologyreport}, a model designed for grounded radiology report generation incorporating lateral views and prior study reports for each subject within the input prompt. \citet{bannur2024maira2groundedradiologyreport} emphasises a positive transfer between this distinct task setup and standard RRG, which falls beyond our study's scope. For comparison and discussion of the latest concurrent and non-LLM-based models, see Appx.~\ref{more-compare}.

Considering these factors, we compared our model with state-of-the-art models, including 
LLaVA-Med \citep{li2023llavamedtraininglargelanguageandvision}, CheXagent \citep{chen2024chexagentfoundationmodelchest}, GPT-4V \citep{openai2024gpt4technicalreport}, Med-PaLM \citep{tu2023generalistbiomedicalai}, LLaVA-Rad and MAIRA-1. Table~\ref{tab:libra-result} presents the results. As many of these models are not publicly available, we present their evaluation results as reported in the original sources.

\subsection{Results}
From Table~\ref{tab:libra-result}, Libra\footnote{Libra was tested on single-image inputs without priors for fair comparison with models lacking temporal modelling.} achieves competitive results across most traditional lexical and clinical metrics, excelling in ROUGE-L, BLEU, METEOR, and RadGraph-based scores. It also leads in the radiologist-aligned RadCliQ metric and CheXbert vector similarity. In the CheXpert classification, it attains the highest Macro-F1 scores and a competitive Micro-F1. Overall, Libra demonstrates robust performance in RRG by effectively leveraging temporal information, with only minor gaps in select clinical metrics. These results highlight the effectiveness of its TAC in capturing temporal contexts and generating clinically relevant radiology reports.

\begin{table*}[t]
    \vspace{-5pt}
    \small
    \centering
    \setlength{\tabcolsep}{4pt}
    \resizebox{0.8\linewidth}{!}{
    \begin{tabular}{l|c|cccc}
        \toprule
        \textbf{Metric} & \textbf{\textit{Libra-1}} & \textbf{w/o TFM} & \textbf{w/o LFE} & \textbf{w/o PIPB} & \textbf{w/o TAC} \\  
        \midrule
        \textbf{Lexical:} & & & & \\
        \hspace{1em} ROUGE-L & 27.56 & 27.33 (-0.85\%) & 27.21 (-1.27\%) & 27.43 (-0.48\%) & 26.17 (-5.04\%) \\
        \hspace{1em} BLEU-1 & 34.84 & 34.17 (-1.92\%) & 34.21 (-1.82\%) & 34.60 (-0.67\%) & 33.03 (-5.20\%) \\
        \hspace{1em} BLEU-4 & 11.51 & 11.13 (-3.33\%) & 11.11 (-3.47\%) & 11.43 (-0.73\%) & 10.02 (-12.98\%) \\
        \hspace{1em} METEOR & 35.50 & 35.06 (-1.24\%) & 34.96 (-1.52\%) & 35.28 (-0.62\%) & 33.98 (-4.28\%) \\ 
        \hspace{1em} BERTScore & 55.87 & 55.60 (-0.49\%) & 55.49 (-0.69\%) & 55.74 (-0.23\%) & 54.63 (-2.22\%) \\
        \midrule
        \rowcolor{lightgray}
        \hspace{1em} F1$_\text{temp}$ & 26.63 & 25.96 (-2.51\%) & 26.21 (-1.57\%) & 26.58 (-0.18\%) & 25.39 (-4.65\%) \\
        \midrule
        \textbf{Clinical:} & & & & \\
        \hspace{1em} RadGraph-F1 & 22.52 & 22.20 (-1.42\%) & 22.03 (-2.19\%) & 22.35 (-0.74\%) & 21.51 (-4.48\%) \\
        \hspace{1em} RG$_{\text{ER}}$ & 27.32 & 26.89 (-1.59\%) & 26.72 (-2.19\%) & 27.09 (-0.84\%) & 25.97 (-4.96\%) \\
        \hspace{1em} RadCliQ$_{\text{0}}$ ($\downarrow$) & 3.10 & 3.12 (-0.65\%) & 3.12 (-0.65\%) & 3.11 (-0.32\%) & 3.15 (-1.61\%) \\
        \hspace{1em} CheXbert vector & 42.02 & 41.57 (-1.07\%) & 41.37 (-1.54\%) & 41.92 (-0.24\%) & 40.93 (-2.59\%) \\
        \hspace{1em} \textit{CheXpert-F1:} & & & & \\
        \hspace{2em} Micro-F1-14 & 52.48 & 51.74 (-1.42\%) & 51.68 (-1.53\%) & 52.13 (-0.67\%) & 51.13 (-2.57\%) \\
        \hspace{2em} Macro-F1-14 & 36.87 & 36.04 (-2.25\%) & 36.12 (-2.03\%) & 36.14 (-1.97\%) & 35.85 (-2.76\%) \\
        \hspace{2em} Micro-F1-5 & 56.63 & 55.37 (-2.23\%) & 55.79 (-1.49\%) & 55.87 (-1.34\%) & 54.51 (-3.74\%) \\
        \hspace{2em} Macro-F1-5 & 49.33 & 47.76 (-3.18\%) & 47.82 (-3.06\%) & 47.98 (-2.75\%) & 47.22 (-4.28\%) \\
        \bottomrule
    \end{tabular}
    }
    \vspace{-4pt}
    \caption{\label{tab:q1}Results of ablation experiments for the Temporal Alignment Connector. `$\downarrow$' indicates that lower is better. Values in (\%) indicate the percentage decrease compared with the \textit{Libra-1}.}
    \vspace{-6pt}
\end{table*}

\section{Ablation Studies}\label{Ablation Study}
We conducted ablation studies on Libra's key components, evaluating module and dataset configurations. All experiments were performed on the MIMIC-CXR test split for the \textit{Findings} generation, with prior images included by default and consistent hyperparameters during training and inference.

\vspace{-5pt}
\paragraph{Does the Temporal Alignment Connector improve model performance?}\label{ablation-q1}
To evaluate the impact of TAC on Libra's performance in RRG, we used a model initialised with the RAD-DINO \citep{pérezgarcía2024raddinoexploringscalablemedical} image encoder, TAC, and Meditron-7b \citep{chen2023meditron70bscalingmedicalpretraining} as the LLM. The baseline (\textit{Libra-1}) was conducted by fine-tuning only the TAC for the \textit{Findings} generation task. As shown in Table~\ref{tab:q1}, we performed ablation studies by progressively removing different TAC components, including the Temporal Fusion Module (TFM), Layerwise Feature Extractor (LFE), Prior Image Prefix Bias (PIPB), and the entire TAC.

Removing TFM restricted the model to single-image processing, akin to LLaVA \citep{liu2023visualinstructiontuning}, but with a four-layer MLP for aligning image features with the LLM's hidden dimensions. Without LFE, the model used the penultimate layer of the encoder. Removing PIPB excluded the mechanism for differentiating true and dummy prior images. Finally, removing the entire TAC left the model reliant solely on the image encoder and LLM.

The results indicate that removing any TAC submodule leads to performance declines across all metrics compared to \textit{Libra-1}. TFM removal caused a notable drop in the F1$_\text{temp}$ score ($\downarrow >$2\%), highlighting its role in capturing temporal information. LFE removal significantly decreased RadGraph-related scores, underscoring its importance in extracting detailed image features. PIPB removal impacted clinical metrics more than lexical metrics, indicating its role in enhancing clinical relevance. Complete TAC removal led to substantial declines in all metrics, demonstrating its critical role in integrating image details and temporal information. The evaluation confirms that TAC plays a vital role in improving Libra's ability to generate high-quality, temporally aware radiology reports. 

For additional ablation studies exploring TAC's contributions, including its impact under general-domain and radiology-specific pre-trained models, its performance after the second training stage, and its robustness through extended fine-tuning and diverse conditions, and an analysis of whether incorporating temporal information improves Libra's performance in RRG tasks, please refer to Appx.~\ref{add-ablation}.

\vspace{-5pt}
\paragraph{Are additional \textit{Impression} and VQA datasets necessary during the feature alignment?}\label{ablation-q2}
To assess the impact of incorporating additional datasets during the first stage of training, we compared a model (\textit{Libra-f}) trained solely on the \textit{Findings} data with Libra, which also used \textit{Impression} and VQA data for feature alignment, as shown in Table~\ref{tab:q2}. 

After the first stage, Libra outperformed \textit{Libra-f} in lexical metrics but showed a slight decline in clinical scores. This decline stems from VQA tasks emphasizing fine-grained, grounded descriptions rather than holistic findings. VQA focuses on individual symptoms, whereas \textit{Findings} integrates multiple normal and abnormal observations, affecting F1$_\text{temp}$ by reducing identified temporal entities.

In the second stage, fine-tuning on \textit{Findings} restored balance, further improving performance. These results indicate that additional datasets enhance Libra's RRG ability, while second-stage fine-tuning ensures well-rounded report generation.

\begin{table}[t]
    \vspace{-6pt} 
    \centering
    \footnotesize
    \setlength{\tabcolsep}{3pt}
    \resizebox{0.89\linewidth}{!}{
    \begin{tabular}{l|cc|cc}
        \toprule
        \multirow{2}{*}{\textbf{Metric}} & \multicolumn{2}{c|}{\textbf{Stage: 1}} & \multicolumn{2}{c}{\textbf{Stage: 2}} \\ 
        \cmidrule(lr){2-3} \cmidrule(lr){4-5}
        & \textit{Libra-f}  & \textbf{Libra}  & \textit{Libra-f}  & \textbf{Libra} \\  
        \midrule
        \textbf{Lexical:} & & & & \\
        \hspace{1em} ROUGE-L & 27.56  & 27.27$^\blacktriangledown$ & 35.31 &  36.66$^\vartriangle$  \\
        \hspace{1em} BLEU-1 & 34.84 & 41.24$^\vartriangle$ & 49.92 & 51.25$^\vartriangle$  \\
        \hspace{1em} BLEU-4 & 11.51 &  13.59$^\vartriangle$ &  23.05 & 24.54$^\vartriangle$ \\
        \hspace{1em} METEOR & 35.50 & 39.44$^\vartriangle$ & 47.99 & 48.90$^\vartriangle$ \\ 
        \hspace{1em} BERTScore & 55.87 & 56.00$^\vartriangle$ &   61.28 & 62.50$^\vartriangle$ \\
        \midrule
        \rowcolor{lightgray}
        \hspace{1em} F1$_\text{temp}$ & 26.63 & 24.80$^\blacktriangledown$   & 33.52 &  35.34$^\vartriangle$ \\
        \midrule
        \textbf{Clinical:} & & & & \\
        \hspace{1em} RadGraph-F1 & 22.52 &  20.45$^\blacktriangledown$  &  30.77 &  32.87$^\vartriangle$ \\
        \hspace{1em} RG$_{\text{ER}}$ &  27.32 & 25.19$^\blacktriangledown$ &  35.44 &  37.27$^\vartriangle$ \\
        \hspace{1em} RadCliQ$_{\text{0}}$ ($\downarrow$) & 3.10 &  3.31$^\blacktriangledown$ &  2.83 &  2.72$^\vartriangle$ \\
        \hspace{1em} CheXbert vector & 42.02 &  35.33$^\blacktriangledown$ &  45.32 &  46.85$^\vartriangle$ \\
        \hspace{1em} \textit{CheXpert-F1:} & & & & \\
        \hspace{2em} Micro-F1-14 & 52.48 &  43.63$^\blacktriangledown$ &  54.11 &  55.87$^\vartriangle$ \\
        \hspace{2em} Macro-F1-14 & 36.87 & 25.68$^\blacktriangledown$ &  37.16 &   40.38$^\vartriangle$ \\
        \hspace{2em} Micro-F1-5 & 56.63 & 49.75$^\blacktriangledown$ & 58.76 &  60.07$^\vartriangle$  \\
        \hspace{2em} Macro-F1-5 & 49.33 &  40.40$^\blacktriangledown$ & 51.99 & 53.75$^\vartriangle$  \\
        \bottomrule
    \end{tabular}
    }
    \vspace{-5pt}
    \caption{\label{tab:q2}Ablation results for dataset configurations. $^\vartriangle$ denotes improvement, while $^\blacktriangledown$ indicates decline.}
    \vspace{-10pt} 
\end{table}

\section{Performance Analysis}\label{Performance Analysis}
We qualitatively assess Libra's ability to generate temporally consistent radiology reports.

\begin{figure}[t]
    \vspace{-6pt}
    \centering
    \includegraphics[width=\columnwidth]{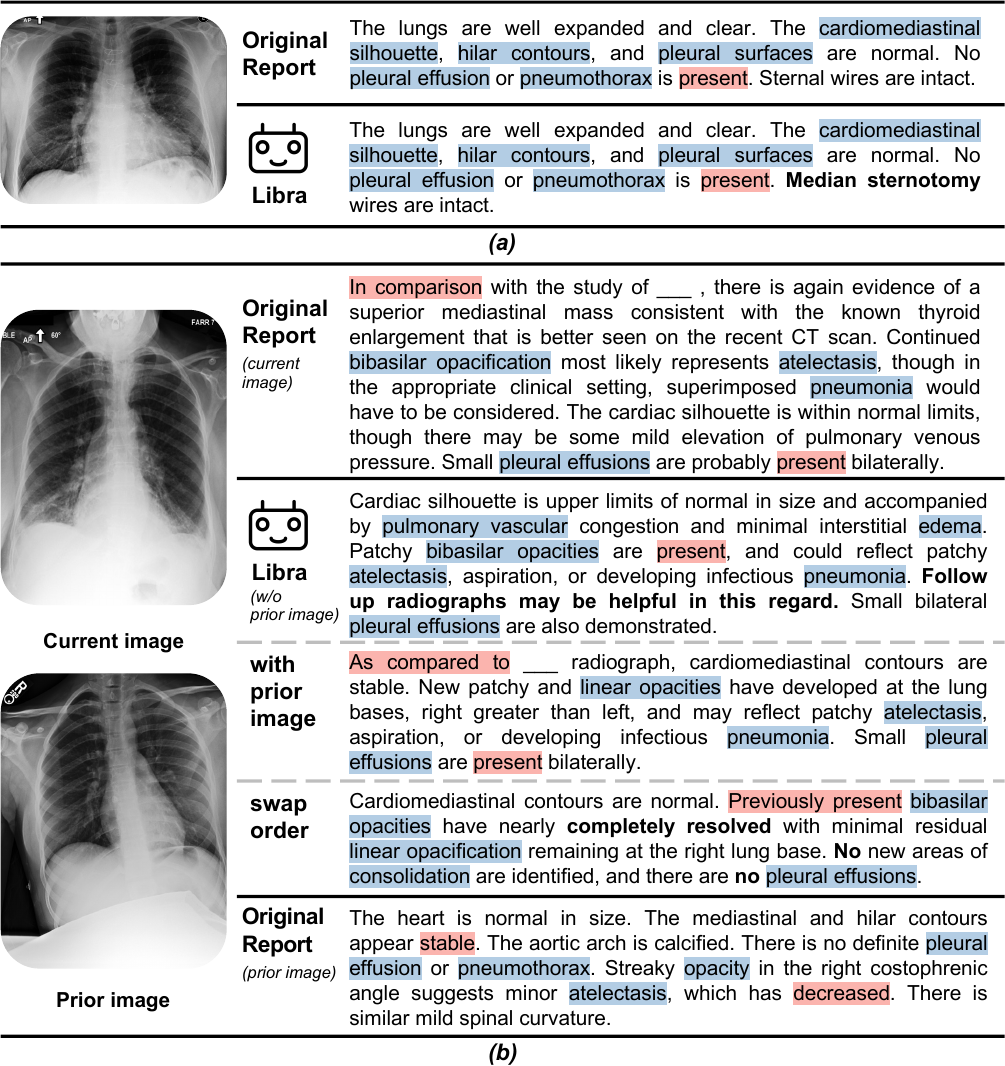}
    \vspace{-20pt}
    \caption{\colorbox{myblue}{Radiological symptoms,}while temporal changes are in \colorbox{myred}{red}. Key highlights presented in \textbf{bold}. Heatmap analysis is available in Appx.~\ref{heatmap-analysis}.}
    \label{fig:examples-section5}
    \vspace{-10pt}
\end{figure}

\vspace{-6pt}
\paragraph{Cases without Prior Image} 
As shown in Figure~\ref{fig:examples-section5} \textbf{\textit{(a)}}, Libra produced detailed descriptions beyond the ground truth, identifying ``sternal wires'' and their type. This demonstrates its capability to deliver clinically relevant information without spurious referencing nonexistent prior studies.

\vspace{-6pt}
\paragraph{Cases with Prior Image} 
In Figure~\ref{fig:examples-section5} \textbf{\textit{(b)}}, new abnormalities such as pleural effusion and pneumonia appeared in the current image. Without a prior image, Libra correctly described the present findings but did not infer disease progression, avoiding spurious references while still suggesting further investigations. When the prior image was considered, Libra effectively captured these progressive changes, provided detailed descriptions, and explicitly referenced the comparison. This facilitated a clear understanding of temporal changes and more accurate descriptions of disease progression.

\vspace{-6pt}
\paragraph{Evaluating Temporal Consistency}\label{TA case}
To assess temporal reasoning, we swapped image order, using the prior image as the current image and vice versa. The generated report then reflected an improved patient condition, aligning with the reversed input sequence but contradicting the ground truth of the original current image. Notably, the report closely resembled the original description of the prior image, as shown at the bottom of Figure~\ref{fig:examples-section5} \textbf{\textit{(b)}}. This indicates that Libra can effectively adapt to both temporal contexts, generating accurate and contextually consistent reports that simulate the conditions of standard clinical practice.

\section{Conclusion}
In this study, we introduced Libra, a temporal-aware multimodal large language model tailored for chest X-ray report generation tasks. Libra employs a two-stage training framework, leveraging a radiology-specific image encoder and language model connected via the Temporal Alignment Connector, enabling seamless integration of visual and textual modalities. Trained solely on the open-access MIMIC-CXR dataset \citep{johnson2019mimic}, Libra demonstrates notable performance gains across key metrics compared to similarly scaled models. Through qualitative and quantitative analysis, we showed that Libra effectively utilises temporal relationships between current and prior scans, addressing challenges such as hallucinations in referencing prior studies. This highlights Libra's ability to generate clinically accurate and temporally consistent radiology reports, setting a new paradigm for multimodal medical AI research. 

Future work will focus on expanding Libra's clinical applicability by incorporating diverse imaging modalities and enhancing temporal reasoning capabilities, and extending it in an agentic way. 

\clearpage
\section*{Limitations}
Despite Libra's ability to model temporal paired images for radiology report generation (RRG), certain limitations remain. First, Libra relies on single prior images for temporal modelling, whereas clinical practice often involves multiple prior scans with varied intervals and angles. Extending the model to handle multiple temporally sequenced images remains an open challenge. Second, our study is based on a single-source dataset with inherent biases in patient demographics and imaging protocols, which may limit generalizability across broader clinical settings. Lastly, while Libra is designed for CXR-based RRG, its applicability to other imaging modalities (e.g., CT, MRI) and integration with structured medical knowledge remains unexplored. For a detailed discussion of these limitations and future directions, see Appx.~\ref{limitations}.

\section*{Ethics Statement}

This work presents Libra, a model designed to enhance radiology report generation by integrating temporal and visual information. While Libra has the potential to improve clinical workflows, reduce radiologist workload, and enhance diagnostic consistency, its deployment must be approached with caution to ensure ethical and responsible use.

Our research exclusively utilises the publicly available and ``de-identified'' MIMIC-CXR dataset \citep{johnson2019mimic}, in accordance with its official documentation, ensuring adherence to ethical and privacy standards under CITI Data or Specimens Only Research certification. By relying solely on open datasets, we prioritise transparency and reproducibility, aligning with best practices in ethical AI research.

This work is intended to support, not replace, medical professionals, ensuring it serves as a complementary tool within clinical practice. While the societal implications are largely positive, further validation across diverse patient populations and healthcare systems is necessary to address potential biases inherent in the dataset. Additionally, it is crucial to mitigate the risks of over-reliance on AI systems, which could inadvertently undermine human oversight or exacerbate healthcare disparities.

Future efforts will aim to extend the model's capabilities to encompass multiple imaging modalities and broader datasets, ensuring greater generalisability, fairness, and adaptability across diverse clinical settings.

\newpage
\bibliography{custom}

\newpage
\appendix

\onecolumn

\section*{Appendix Contents} 
\startcontents[sections]
\printcontents[sections]{l}{1}{%
    \setcounter{tocdepth}{2} 
}

\newpage
\section{Related Work}\label{related-work}
\subsection{Radiology Report Generation} 
Radiology report generation (RRG) aims to address the long-tail distribution of observations in chest X-rays (CXRs) and produce fine-grained descriptions of clinical findings, making it a key objective in automated medical imaging analysis \citep{wang2018tienettextimageembeddingnetwork}.

Early RRG systems relied on recurrent neural networks (RNNs) \citep{liu2019clinicallyaccuratechestxray}, which have since been largely replaced by transformer-based architectures \citep{miura2021improvingfactualcompletenessconsistency, chen2022generatingradiologyreportsmemorydriven}, including large language models (LLMs) such as PaLM \citep{chowdhery2022palmscalinglanguagemodeling} and Vicuna-7B \citep{vicuna2023}. These models excel at language generation, offering substantial improvements in fluency and factual accuracy.

To further enhance clinical accuracy, some methods incorporate reinforcement learning (RL) to optimise for task-specific rewards, such as capturing ``clinically relevant'' features \citep{liu2019clinicallyaccuratechestxray,irvin2019chexpertlargechestradiograph} or maintaining logical consistency \citep{miura-etal-2021-improving,delbrouck-etal-2022-improving}. However, these approaches often rely on external tools like CheXbert \citep{smit-etal-2020-combining} or RadGraph \citep{jain2021radgraphextractingclinicalentities}, adding complexity to the optimisation process.

Recent advancements in LLMs have shown that plain auto-regressive language modelling can achieve strong performance in RRG tasks. However, RL-based objectives and task-specific optimisations remain complementary, offering additional opportunities for improvement. Research on leveraging temporal information in RRG tasks can be broadly categorised into LLM-based and non-LLM-based methods, each presenting distinct advantages and challenges.

\subsection{LLM-based Model}\label{LLM-based-Model}
LLM-based models have achieved significant success in the RRG task, primarily due to advancements in visual instruction tuning \citep{liu2023visualinstructiontuning}. Structurally, these models \citep{li2023llavamedtraininglargelanguageandvision,chaves2024clinicallyaccessibleradiologyfoundation,hyland2024maira1specialisedlargemultimodal,zhou2024generalistlearnermultifacetedmedical,park2024m4cxrexploringmultitaskpotentials} typically consist of an image encoder and an adapter that connects the encoder's outputs to the LLM. Most existing adapters use single-layer hidden representations (e.g., the last or penultimate layer) from pre-trained image encoders, limiting their ability to integrate features from multiple images effectively.

In end-to-end training, LLM-based models handle multiple image inputs by concatenating them with textual prompts, forming a composite input to the LLM. For instance, the input format is often structured as ``\verb|<|Current Image Placeholder\verb|>| + \verb|<|Prior Image Placeholder\verb|>| + \verb|<|Prompt\verb|>|''. However, this approach provides limited guidance on the relationship between the images within the prompt. \citet{Din_HiA_MICCAI2024} proposed the High-Resolution Instruction-Aware Adapter (HiA) to refine image-text representations, improving the model's ability to follow textual prompts with multiple images. While this enhances instruction adherence, it does not explicitly model relationships between paired images.

In contrast to this vanilla approach, Libra explicitly models temporal relationships in paired images through its Temporal Alignment Connector (TAC). Instead of simply concatenating images in the LLM's latent space, TAC leverages all hidden-layer features from the image encoder to provide richer feature representations. By directly modelling temporal dynamics, Libra enables more precise and context-aware radiology report generation.

\subsection{non-LLM-based Model} 
Non-LLM-based models typically employ transformer encoder-decoder architectures or their variants, which often require separate training for individual modules. These approaches handle ``single-'' and ``double-'' image inputs by symbolically differentiating tasks and employing distinct architectures tailored for each input type. Additionally, they frequently incorporate extra information such as prior reports, symptom labels, and knowledge graphs.

For instance, \citet{serra2023controllablechestxrayreport} uses symbolic alignment in its Longitudinal Projection Module along with a separately trained BERT-based \citep{devlin2019bertpretrainingdeepbidirectional} text generator. RECAP \citep{hou-etal-2023-recap} implements a two-stage training process: classification tasks followed by report generation, leveraging a transformer encoder-decoder with symbolic task differentiation. TiBiX \citep{sanjeev2024tibixleveragingtemporalinformation} incorporates causal attention layers and learnable padding tokens to handle cases without prior images, while BioViL-T \citep{bannur2023learningexploittemporalstructure} is a self-supervised vision-language training framework that features a CNN–Transformer hybrid multi-image encoder trained jointly with a BERT-based text model.

On one hand, the difference in model parameter sizes, and on the other, as LLM-based models generally outperform other types of models (i.e. non-LLM-based) in the RRG task, papers on non-LLM-based models or those using small language models (SLMs) typically do not compare their methods with LLM-based approaches. Nonetheless, we conducted comparisons and discussions to reaffirm this observation, as detailed in Appx.~\ref{non-llm}.

\subsection{Radiological Image Representation}

Radiology-specific pre-trained image encoder models are essential for RRG tasks due to the unique characteristics of radiological images, which fall outside the distribution of general-domain image models \citep{pérezgarcía2024raddinoexploringscalablemedical}.

Several notable advancements have been made in this domain. \citet{zhou2023advancingradiographrepresentationlearning} proposed Masked Record Modeling (MRM), a unified framework combining self-supervision with radiology report supervision to enhance radiograph representation learning. Similarly, BioViL-T \citep{bannur2023learningexploittemporalstructure} employs a CNN-Transformer hybrid architecture to model multimodal relationships and leverage temporal structures for tasks such as disease progression classification and report generation. In addition, BiomedCLIP \citep{zhang2024biomedclipmultimodalbiomedicalfoundation} is a multimodal biomedical foundational model pre-trained across diverse biomedical tasks. 

RAD-DINO \citep{pérezgarcía2024raddinoexploringscalablemedical} is a medical image encoder that employs a pure image-based self-supervised learning approach from DINOv2 \citep{oquab2024dinov2learningrobustvisual} for continuous pretraining, focusing exclusively on image data to avoid the limitations of text supervision. Recent works have extensively applied RAD-DINO to RRG tasks, including MAIRA-2 \citep{bannur2024maira2groundedradiologyreport} and M4CXR \citep{park2024m4cxrexploringmultitaskpotentials}. Notably, \citet{pérezgarcía2024raddinoexploringscalablemedical} demonstrated that RAD-DINO outperforms other image encoders in RRG tasks. 

Building on this evidence, our model incorporates RAD-DINO as its image encoder to ensure high-quality radiological image representations, providing a robust foundation for downstream RRG tasks.

\section{Research Objectives}\label{research-object}

\subsection{Temporal Information}\label{research-object-ti}
Temporal changes are critical for understanding disease progression. In radiology, paired images and their corresponding reports document subtle evolutions of symptoms over time. This temporal information is often captured by comparing current scans with prior ones to highlight symptom evolution or newly identified findings.

The relative positioning of scans within the timeline determines the extent of temporal information. Therefore, the relative timing between scans is key: when the prior scan is recent, reported changes tend to be minimal; conversely, an older prior scan reveals more pronounced differences. 

Importantly, while temporal context enriches the diagnostic narrative, it does not alter the factual observations present in the current scan—it merely provides additional layers of interpretative insight.

\subsection{Research Aims}
This study aims to enhance radiology report generation (RRG) by effectively incorporating temporal information into the modelling process. In clinical practice, chest X-ray (CXR) analysis often depends on comparing the current scan with the prior image to capture disease progression and evolution. Our primary objective is to leverage these temporal cues to generate more accurate, context-aware radiological reports that faithfully reflect both stable conditions and clinically significant changes.

Unlike previous LLM-based models (discussed in Appx.~\ref{LLM-based-Model}), which depend on the LLM to infer temporal information solely from text, our approach explicitly models temporal relationships at the architectural level. Inspired by the principle \textbf{``structure determines function''} \citep{fischer1894einfluss,doi:10.1073/pnas.37.4.205,watson1953molecular}, we introduce the Temporal Alignment Connector (TAC), a dedicated module designed to capture temporal dynamics. Details are provided in Sec.~\ref{Temporal Alignment Connector}.

\subsection{Research Scope}
This study focuses on frontal chest X-rays, treating each examination per image while incorporating a single prior image as an auxiliary input when available. Rather than modelling patient-level longitudinal history, our goal is to generate a report for the current image while leveraging temporal information from one preceding scan. To ensure fairness in benchmarking, Libra was evaluated on single-image inputs without priors (see Table~\ref{tab:libra-result}). Yet, temporal information remains implicitly present through several factors:

\begin{itemize}[align=right,itemindent=2em,labelsep=2pt,labelwidth=1em,leftmargin=0pt,nosep,topsep=2pt]
    \setlength{\itemsep}{2pt}
    \item Explicit temporal states (e.g., ``stable'' or ``unstable'') are frequently described in radiology reports.
    \item Latent temporal progression exists in the dataset, as prior studies influence diagnostic phrasing.
    \item The absence of a prior image itself constitutes a temporal scenario, representing an extreme case where the patient's condition is assumed stable due to a lack of comparative reference.
\end{itemize}

Our model can effectively handle scenarios with limited temporal information in the RRG task. For instance, in a case where a patient has two scans taken just milliseconds apart, the current and prior images would be nearly identical, as no pathological changes would manifest within such a short interval. This extreme scenario demonstrates how the model handles clinical practice under limited temporal information. In such cases, the correct diagnosis for this minimal interval would be that the patient's condition is ``stable''; our model should then generate a report reflecting this stability. When no prior image is available, we employ a dummy prior image (a copy of the current image) to maintain input consistency and mitigate spurious references to nonexistent priors.

However, in clinical practice, patients often undergo multiple prior scans, sometimes from different orientations, providing a more complex temporal context. This lies beyond the scope of our current study, and a detailed discussion of such scenarios is provided in Appx.~\ref{limitations}.

\section{Prompt Example}\label{prompt-example}

\begin{table}[h]
    \vspace{-15pt}
    \centering
    \small
    \begin{tabular}{p{0.9\linewidth}} 
        \toprule
        \textbf{Original Radiology Report} \\
        \midrule
        EXAMINATION: Chest (Portable AP) \\
        INDICATION: Dyspnea and cough, right-sided back pain. \\
        HISTORY: Intubation with pulmonary edema. \\
        COMPARISON: Chest radiographs on \_\_\_ and CT chest without contrast on \_\_\_. \\
        TECHNIQUE: Portable upright chest radiograph. \\
        FINDINGS: In comparison with the prior study, there are diffuse bilateral pulmonary opacifications, more prominent on the right. These findings could indicate severe pulmonary edema, but superimposed pneumonia or developing ARDS cannot be excluded. Monitoring and support devices are appropriately positioned. \\
        \bottomrule
        \toprule
        \textbf{Prompt Content} \\
        \midrule
        \textcolor{red}{\textit{[System prompt]:}} \textbf{\{} \\
        The assistant specialised in comparing Chest X-ray images, identifying differences, and noting temporal changes.\\
        \textbf{\}} \\
        \textbf{$+$}\\
        \textit{$<$Image Representation Placeholder$>$} \\
        \textbf{$+$}\\
        \textcolor{blue}{\textit{[Clinical prompt]:}} \textbf{\{} \\
        Provide a detailed description of the findings in the radiology image. Following clinical context: \\
        Indication: Dyspnea and cough, right-sided back pain. \\
        History: Intubation with pulmonary edema. \\
        Comparison: Chest radiographs on \_\_\_ and CT chest without contrast on \_\_\_. \\
        Technique: Portable upright chest radiograph. \\
        \textbf{\}} \\
        \bottomrule
    \end{tabular}
    \vspace{-5pt}
    \caption{\label{tab:promptexample}Examples of Libra's system and clinical prompts for \textit{Findings} section generation in the RRG task.}
    \vspace{-10pt}
\end{table}

We selected examples from the MIMIC-CXR \citep{johnson2019mimic} dataset and synthesised them using GPT-4 \citep{openai2024gpt4technicalreport} to ensure ethical compliance, as illustrated in Table~\ref{tab:promptexample}. Following the rule-based approach by \citet{hyland2024maira1specialisedlargemultimodal}, we extracted key sections from the report of the current image. Each example combines a fixed system prompt with a dynamic clinical prompt tailored to the current scan. We utilised four clinical instructions from the original report: \{\textit{Indication}\}, \{\textit{History}\}, \{\textit{Comparison}\}, and \{\textit{Technique}\}. In contrast, MAIRA-2 \citep{bannur2024maira2groundedradiologyreport}, which incorporates prior image reports, our approach focuses exclusively on the current image's context, maintaining a clear distinction from prior study information of the report.

\section{Training Configuration}\label{train-config}

Libra is trained using a standard auto-regressive language modelling loss (cross-entropy). For this study, we employ Meditron-7b \citep{chen2023meditron70bscalingmedicalpretraining} as the LLM, with a total batch size of 16 throughout the training process. The training is conducted on a computational infrastructure equipped with A6000 GPU (48GB of memory) and using DeepSpeed optimization \citep{rajbhandari2020zeromemoryoptimizationstraining} with ZeRO-2 for stage 1 and ZeRO-3 for stage 2, and BF16 precision is enabled.

A cosine learning rate scheduler is employed, starting with a warm-up phase of $0.03$. In the first stage of training, we run for 1 epoch ($\sim$385 hours) with a learning rate of $2 \times 10^{-5}$. In the second stage, the model is trained for 3 epochs ($\sim$213 hours) at the same learning rate. The LoRA \citep{hu2021loralowrankadaptationlarge} parameters are set to $r = 128$ and $alpha = 256$. The final checkpoint for all runs is selected based on the observation of the minimum loss on the evaluation dataset throughout the training process.

\textbf{Note:} Prior work, especially in the medical domain, typically employs full model fine-tuning for RRG tasks. However, due to hardware constraints, we can only adopt a lightweight training technique for parameter-efficient adaptation. As a result, our approach may underperform full model fine-tuning strategies in the second stage, despite maintaining computational efficiency.

\section{Datasets Description}\label{detail-dataset}

\begin{table*}[ht]
    \vspace{-5pt}
    \centering
    \small
    \setlength{\tabcolsep}{3pt} 
    \begin{tabular}{ll|ccc|cccc}
        \toprule
        \textbf{Dataset} & \textbf{Task Type} & \multicolumn{3}{c|}{\textbf{\# Samples}}   & \multicolumn{3}{c}{\textbf{\% Has Prior}} \\
        & &  \textbf{Train (\%)} & \textbf{Valid (\%)} & \textbf{Test (\%)}  & \textbf{Train} & \textbf{Valid} & \textbf{Test} \\
        \midrule
        MIMIC-CXR  & Findings & 162 955 (13.43\%) & 1286 (0.88\%) & 2461 (2.78\%)  & 58.43 & 60.11 & 86.03 \\
         & Impression  & 199 548 (16.45\%)& 1671 (1.14\%) & 2343 (2.64\%)  & 64.85 & 67.09 & 85.49 \\
        \midrule
        
        Medical-Diff-VQA & Difference & 131 563 (10.85\%)  & 16 372 (11.17\%) & 16 389 (18.48\%) & 100 & 100 & 100 \\
        & Abnormality  & 116 394 (9.59\%) & 14 512 (9.90\%) & 14 515 (16.37\%) & 100 & 100 & 100 \\
        & Presence  & 124 654 (10.28\%) & 15 549 (10.61\%) & 15 523 (17.51\%) & 100 & 100 & 100 \\
        & View  & 44 970 (3.71\%) & 5696 (3.89\%) & 5599 (6.31\%) & 100 & 100 & 100 \\
        & Location  & 67 187 (5.54\%) & 8510 (5.81\%) & 8496 (9.58\%) & 100 & 100 & 100 \\
        & Level  & 53 728 (4.43\%) & 6722 (4.59\%) & 6846 (7.72\%) & 100 & 100 & 100 \\
        & Type  & 22 067 (1.82\%) & 2709 (1.85\%) & 2702 (3.05\%) & 100 & 100 & 100 \\
        \midrule
        MIMIC-Ext-\textit{MIMIC} & Presence & 109 455 (9.02\%) & 26 153 (17.84\%) & 4566 (5.15\%) & 0 & 0 & 0 \\
        \textit{-CXR-VQA} & Anatomy & 37 952 (3.13\%) & 10 210 (6.96\%) & 1963 (2.21\%) & 0 & 0 & 0 \\
        & Attribute & 49 948 (4.12\%) & 13 111 (8.94\%) & 2578 (2.91\%) & 0 & 0 & 0 \\
        & Abnormality & 60 692 (5.00\%) & 16 109 (10.99\%) & 3199 (3.61\%) & 0 & 0 & 0 \\
        & Size & 16 000 (1.32\%) & 4000 (2.73\%) & 705 (0.80\%) & 0 & 0 & 0 \\
        & Plane & 7992 (0.66\%) & 1992 (1.36\%) & 386 (0.44\%) & 0 & 0 & 0 \\
        & Gender & 7992 (0.66\%) & 1992 (1.36\%) & 396 (0.45\%) & 0 & 0 & 0 \\
        
        \midrule
        \textbf{Total}  & Multi-type & 1 213 097 (100\%) & 146 594 (100\%) & 88 669 (100\%) & 64.73 & 49.09 & 83.67 \\
        \bottomrule
    \end{tabular}
    \vspace{-5pt}
    \caption{\label{tab:datasets}Datasets used for training and evaluating Libra include statistics on the proportion of samples that contain prior images. The first stage uses the full dataset, while the second stage fine-tunes for downstream tasks.}
    \vspace{-10pt}
\end{table*}

\paragraph{MIMIC-CXR} \citep{johnson2019mimic}\quad This is a large, publicly accessible dataset comprising 377,110 DICOM images across 227,835 studies, each accompanied by a radiology report \citep{johnson2019mimic}. For images, we use the commonly available JPEG files from MIMIC-CXR-JPG \citep{johnson2019mimiccxrjpglargepubliclyavailable}, rather than the original DICOM files, and we preprocess the dataset to exclude non-AP/PA scans. For each report, we extract the \textit{Findings, Impression, Indication, History, Comparison,} and \textit{Technique} sections using rule-based heuristics supported by the official MIMIC code repository \citep{johnson2018mimic}.

For the \textit{Findings} section generation task, studies without extractable \textit{Findings} are discarded, while other missing sections are permitted. The same approach is applied to the \textit{Impression} section generation task. In all our experiments, we adhere to the official MIMIC-CXR dataset split. 

Meanwhile, we retrieve prior images by following the chronological order of studies as indicated by the official labels, selecting the closest prior study as the reference image. It is important to note that, to prevent data leakage between the train, validation, and test sets, prior images are retrieved only from within the same split.

\paragraph{Medical-Diff-VQA} \citep{10.1145/3580305.3599819}\quad This dataset is a derivative of the MIMIC-CXR dataset, focused on identifying differences between pairs of main and reference images. The data split adheres to the original labelling, ensuring no data leakage occurs. In total, this dataset comprises 700,703 question-answer pairs derived from 164,324 main-reference image pairs. As shown in Table~\ref{tab:datasets}, the questions are divided into seven categories: abnormality, location, type, view, presence, and difference. 

Each pair consists of a main (current) image and a reference (prior) image, both taken from different studies of the same patient. The reference image is always selected from an earlier visit, with the main image representing the later visit. Of the seven question types, the first six types focus on the main image, while the ``difference'' questions involve both images.

\paragraph{MIMIC-Ext-\textit{MIMIC-CXR-VQA}} \citep{bae2023ehrxqamultimodalquestionanswering}\quad This dataset extends MIMIC-CXR for VQA tasks tailored to CXRs. It includes questions generated from 48 unique templates covering seven content types: presence, anatomy, attribute, abnormality, size, plane, and gender, as shown in Table~\ref{tab:datasets}. Each template was developed with the guidance of board-certified medical experts to ensure clinical relevance, addressing both standard medical VQA content and more complex logical scenarios. In total, the dataset consists of 377,391 unique entries. Since annotations are based on single images, the current image serves as a dummy prior image for all entries in our experiment.
\vspace{5pt}
\\
For this study, we carefully selected datasets that provide complete reports and temporal information (i.e., prior images) to ensure alignment with our research objectives (see Appx.~\ref{research-object}) for the RRG task. After thoroughly evaluating other datasets, we found them \textbf{unsuitable} for the following reasons:

\paragraph{CheXpert} \citep{irvin2019chexpertlargechestradiograph}\quad This dataset includes annotated scans with label-specific annotations rather than full medical reports. While useful for training image encoders or annotation models, it is not appropriate for the RRG task, which requires complete diagnostic reports.

\paragraph{PadChest} \citep{bustos2020padchest}\quad Although it includes reports and corresponding prior images, its reports are in Spanish, placing cross-language training beyond the scope of our model.

\paragraph{IU-Xray} \citep{demner2016preparing}\quad This dataset lacks patient-level metadata and prior study information, which is critical for our focus on temporal information in chest X-rays.

\paragraph{Chest ImaGenome Dataset} \citep{wu2021chestimagenomedatasetclinical}\quad Although derived from MIMIC-CXR \citep{johnson2019mimic}, it does not follow the official split, raising concerns about potential data leakage between training, validation, and test sets.
\vspace{5pt}
\\
Meanwhile, the following two datasets were processed using GPT-4 \citep{openai2024gpt4technicalreport} to eliminate hallucinated references to prior exams. While this prevents erroneous comparisons, it also removes essential temporal information originally present in the reports, potentially affecting tasks that rely on temporal reasoning.

\paragraph{LLaVA-Rad MIMIC-CXR Dataset} \citep{chaves2024clinicallyaccessibleradiologyfoundation}\quad This dataset was refined using GPT-4 \citep{openai2024gpt4technicalreport} through a structured text-cleaning pipeline. The process involved: (1) correcting typographical errors and split words, (2) removing redundant or repeated phrases to improve clarity, (3) eliminating explicit temporal references (e.g., “Compared to the prior study, no significant interval change was noted”) to ensure the report focuses exclusively on the current image, and (4) restructuring content into standardised sections, including \textit{Indication}, \textit{Findings}, and \textit{Impression}.

\paragraph{ReXPref-Prior Dataset} \citep{banerjee2024directpreferenceoptimizationsuppressing}\quad A modified version of MIMIC-CXR \citep{johnson2019mimic} in which GPT-4 \citep{openai2024gpt4technicalreport} systematically removes all references to prior exams from both the \textit{Findings} and \textit{Impression} sections. While this adjustment prevents spurious prior-study references, it also eliminates crucial temporal context, limiting its suitability for applications requiring longitudinal assessment of disease progression.

\section{Evaluation Metrics}\label{Metrics}

\subsection{Lexical Metrics}

We employed standard natural language generation metrics to quantify the overlap between generated and reference reports. Specifically, ROUGE-L \citep{lin-2004-rouge} measures the length of the longest common subsequence between the generated and reference reports. BLEU-$\{1,4\}$ \citep{10.3115/1073083.1073135} calculates n-gram precision and applies a brevity penalty to discourage overly short predictions. METEOR \citep{banerjee-lavie-2005-meteor}, computes the weighted harmonic mean of unigram precision and recall, with an additional penalty for fragmenting consecutive word sequences. Finally, we report BERTScore \citep{zhang2020bertscoreevaluatingtextgeneration}, which leverages pre-trained contextual embeddings from BERT \citep{devlin2019bertpretrainingdeepbidirectional} to match words in candidate and reference sentences based on cosine similarity. We used default parameters for all of these evaluation metrics.

\subsection{Clinical Metrics}

For radiology-specific metrics, we used as many of the same evaluation scores as possible from previous studies \citep{tu2023generalistbiomedicalai,hyland2024maira1specialisedlargemultimodal,bannur2024maira2groundedradiologyreport,chaves2024clinicallyaccessibleradiologyfoundation}, including the following:

\paragraph{RadGraph-based metrics} 
RadGraph model \citep{jain2021radgraphextractingclinicalentities} is designed to parse radiology reports into structured graphs. These graphs consist of clinical entities, which include references to anatomy and observations, as well as the relationships between these entities. This structured representation enables a more detailed and systematic analysis of radiology reports, facilitating downstream tasks such as information extraction, report generation, and clinical decision support. 

These include RadGraph-F1 \citep{jain2021radgraphextractingclinicalentities}, which computes the overlap in entities and relations separately and then reports their average. And a variant of it, RG$_{\text{ER}}$ \citep{delbrouck2022improvingfactualcorrectnessradiology}, which matches entities based on their text, type, and whether they have at least one relation\footnote{RG$_{\text{ER}}$ is implemented as \texttt{F1RadGraph} with \texttt{reward=partial} by the radgraph package.}.

\paragraph{CheXpert F1} This set of metrics utilizes the CheXbert automatic labeler \citep{smit-etal-2020-combining} to extract ``present'', ``absent'', or ``uncertain'' labels for each of the 14 CheXpert pathologies \citep{irvin2019chexpertlargechestradiograph} from the generated reports and their corresponding references. In line with prior work, we report CheXpert-F1 for all 14 classes, as well as for the 5 most common findings in real-world CXR reports, referring to these as ``[Macro/Micro]-F1-[5/14]''.

\paragraph{CheXbert vector similarity} We also employ CheXbert vector similarity \citep{Yu2022.08.30.22279318}, which calculates the cosine similarity between the embeddings of the generated and reference reports after processing them through the CheXbert model \citep{smit-etal-2020-combining}. 

\paragraph{RadCliQ} In addition, we utilise RadCliQ (Radiology Report Clinical Quality)~\citep{Yu2022.08.30.22279318}, a composite metric that combines RadGraph-F1 and BLEU scores in a linear regression model to estimate the number of errors that radiologists are likely to detect in a report. To maintain consistency with previous research, we use version $0$ of it. 

Both the CheXbert vector similarity, RadCliQ$_{\text{0}}$, and RadGraph-F1 metrics are calculated using the code released by \citet{Yu2022.08.30.22279318}.

\subsection{Temporal Entity F1}
We introduced $F1_\text{temp}$, a metric specifically designed to detect temporal entities reflecting changes over time. Unlike traditional lexical or radiology-specific metrics, $F1_\text{temp}$ evaluates the quality of temporal information in radiology reports.

\begin{table}[ht]
    \centering
    \small
    \vfill
    \renewcommand{\arraystretch}{1.3} 
    \begin{tabular}{p{2.2cm}|p{7cm}|ccc}
        \toprule
        \textbf{Ground Truth} & \textbf{Candidate} & ROUGE-L & RadGraph-F1 & $F1_\text{temp}$ \\
        \midrule
        \multirow{2}{2cm}{Compare with prior scan, pleural effusion has\colorbox{myred}{worsened.}} 
        & The pleural effusion has progressively\colorbox{myred}{worsened}since previous scan. 
        & 0.47 & 0.86 & \textbf{1.0} \\
        \cmidrule(lr){2-5}
        & The pleural effusion is noted again on the current scan. 
        & 0.22 & 0.80 & \textbf{0.0} \\
        \bottomrule
    \end{tabular}
    \vspace{-5pt}
    \caption{\label{tab:temporal-f1-example}Evaluation of candidate reports using the Temporal Entity F1 score ($F1_\text{temp}$). Descriptions of temporal changes are\colorbox{myred}{marked.}}
    \vspace{-10pt}
\end{table}

As shown in Table~\ref{tab:temporal-f1-example}, the differences in lexical (ROUGE-L \citep{lin-2004-rouge}) and clinical (RadGraph-F1 \citep{jain2021radgraphextractingclinicalentities}) metrics between the two candidates are relatively smaller compared to the $F1_\text{temp}$ score. This demonstrates that Temporal Entity F1 effectively captures and evaluates the quality of temporal information in radiology reports, distinguishing it more accurately than other standard metrics in the context of temporal information descriptions.

\section{Analysis of Concurrent Work and Non-LLM-based Models}\label{more-compare}

\subsection{Discussion on Performance with Radiology Foundation Models}

\begin{table}[ht]
    \begin{center}
    \small
    \begin{tabular}{l|cccc|c}
        \toprule
        \textbf{Metric} & \textbf{RaDialog} & \textbf{MedVersa} & \textbf{MAIRA-2} & \textbf{M4CXR} & \textbf{Libra (\%)} \\  
        \midrule
        \textbf{Lexical:} & & & & \\
        \hspace{1em} ROUGE-L & 31.6 & -- & \textbf{38.4} & 28.5 & \underline{36.7} (-4.4\%)\\
        \hspace{1em} BLEU-1 & 39.2 & -- & \underline{46.5}  & 33.9 & \textbf{51.3} (10.3\%)\\
        \hspace{1em} BLEU-4 & 14.8 & 17.8  & \underline{23.4}  & 10.3 & \textbf{24.5} (4.7\%)\\
        \hspace{1em} METEOR & -- & -- & \underline{42.0}  & -- & \textbf{48.9} (16.4\%)\\ 
        \hspace{1em} BERTScore & -- & \underline{49.7}  & -- & -- & \textbf{62.5} (25.8\%)\\
        \midrule
        \textbf{Clinical:} & & & & \\
        \hspace{1em} RadGraph-F1 & -- & 28.0 & \textbf{34.6}  & 21.8 & \underline{32.9} (-4.9\%)\\
        \hspace{1em} RG$_{\text{ER}}$ & -- & -- & \textbf{39.7}  & --  & \underline{37.6} (-5.3\%)\\
        \hspace{1em} RadCliQ$_{\text{0}}$($\downarrow$)& --  & \underline{2.7}  & \textbf{2.6}  & --  & \underline{2.7} (-3.8\%)\\
        \hspace{1em} CheXbert vector& --  & 46.4   & \textbf{50.6}  & -- & \underline{46.9} (-7.3\%)\\
        \hspace{1em} \textit{CheXpert-F1:} & & & & \\
        \hspace{2em} Micro-F1-14 & 39.2 & -- & \underline{58.5}  & \textbf{60.6}   & 55.9 (-7.8\%)\\
        \hspace{2em} Macro-F1-14 & --  & -- & \textbf{42.7}  & 40.0  & \underline{40.4} (-5.4\%)\\
        \hspace{2em} Micro-F1-5 & --   & -- & 58.9  & \textbf{61.8}  & \underline{60.1} (-2.8\%)\\
        \hspace{2em} Macro-F1-5& --  & -- & \underline{51.5}  & 49.5   & \textbf{53.8} (4.5\%)\\  
        \bottomrule
    \end{tabular}
    \vspace{-5pt}
    \caption{\label{tab:appendixd}Findings section generation performance of Libra and the latest concurrent work. The best performances are highlighted in \textbf{bold}, and the second-best scores are \underline{underlined}. `$\downarrow$' indicates that lower values are better. `--' indicates missing data. The percentage ($\%$) indicates the improvement over the best existing model.}
    \end{center}
    \vspace{-15pt}
\end{table}

As shown in Table~\ref{tab:appendixd}, these models belong to the category of radiology foundation models. 

DaDialog \citep{pellegrini2023radialoglargevisionlanguagemodel} is a conversational MLLM designed for a broad range of dialogue-based medical assistance tasks. To enhance structured findings extraction, it employs the publicly available CheXbert model \citep{smit2020chexbert} to extract symptom labels from scans, facilitating a structured representation of findings.

MedVersa \citep{zhou2024generalistlearnermultifacetedmedical} and M4CXR \citep{park2024m4cxrexploringmultitaskpotentials} support a diverse set of tasks, including medical report generation, visual grounding, and visual question answering. These models aim to provide general-purpose multimodal medical assistance by leveraging vision-language pre-training strategies.

MAIRA-2 \citep{bannur2024maira2groundedradiologyreport} specialises in grounded radiology report generation, which differs from traditional report generation tasks by requiring explicit image-level localization of findings and symptoms. Grounded radiology reporting, as defined by \citet{bannur2024maira2groundedradiologyreport}, structures the report as a list of sentences, where each sentence: (1) is linked to zero or more spatial image annotations, and (2) describes at most a single finding from an image. To support this task, MAIRA-2 introduces a custom dataset, explicitly designed to provide structured annotations aligning textual descriptions with spatial regions of interest in radiological images. This approach contrasts with conventional RRG models that generate unstructured free-text reports.

It is worth noting that the inference sets differ slightly across these models. Additionally, all these models leverage supplementary radiology information, such as lateral view scans, prior study reports, or both (as detailed in Appx.~\ref{LLM-based-Model}), to enhance their performance in radiology-related tasks.

Despite these considerations, Libra achieves the highest scores on most lexical metrics, including BLEU-$\{1,4\}$, METEOR, and BERTScore, while trailing slightly behind MAIRA-2 on ROUGE-L. In clinical metrics, Libra predominantly ranks second, just behind the best-performing model. For clinical metrics, Libra consistently ranks second, just behind the top-performing model. In metrics that evaluate medical entities and their relationships, such as RadGraph-F1, RG$_{\text{ER}}$, and RadCliQ, Libra also ranks second. Similarly, Libra comes second in the CheXbert vector embedding score. However, in the CheXpert metrics, Libra ranks first in Macro-F1 for the 5-class subset, with only a slight dip in the Micro-F1 score for the 14-class subset.

Incorporating lateral images and prior study reports could enhance clinical scores. Additionally, strategies like chain-of-thought reasoning and grounded report generation further improve performance in RRG tasks. Looking ahead, we plan to develop model architectures that can automatically adapt to multiple tasks and diverse scenarios, enabling more efficient handling of additional radiological information.

\subsection{Discussion on Performance with non-LLM-based Models}\label{non-llm}

\begin{table}[ht]
    \centering
    \small
    \begin{tabular}{l|cccccc|ccc}
        \toprule
        \multirow{2}{*}{\textbf{Model}} & \multicolumn{6}{c|}{\textbf{Lexical Metrics}} & \multicolumn{3}{c}{\textbf{Clinical Metrics}} \\
        \cmidrule(lr){2-7} \cmidrule(lr){8-10}
         & \textbf{B-1} & \textbf{B-2} & \textbf{B-3} & \textbf{B-4} & \textbf{MTR} & \textbf{R-L} & \textbf{P} & \textbf{R} & \textbf{F$_1$} \\
        \midrule
        ST$^\text{‡}$ & 29.9 & 18.4 & 12.1 & 8.4 & 12.4 & 26.3 & 24.9 & 20.3 & 20.4 \\
        ATT2IN$^\text{‡}$ & 32.5 & 20.3 & 13.6 & 9.6 & 13.4 & 27.6 & 32.3 & 23.9 & 20.4 \\
        ADAATT$^\text{‡}$ & 29.9 & 18.5 & 12.4 & 8.8 & 11.8 & 26.6 & 26.8 & 18.6 & 18.1 \\
        TopDown$^\text{‡}$ & 31.7 & 19.5 & 13.0 & 9.2 & 12.8 & 26.7 & 32.0 & 23.1 & 23.8 \\
        R2Gen & 35.3 & 21.8 & 14.5 & 10.3 & 14.2 & 27.0 & 33.3 & 27.3 & 27.6 \\
        R2GenCMN & 35.3 & 21.8 & 14.8 & 10.6 & 14.2 & 27.8 & 34.4 & 27.5 & 27.8 \\
        XPRONET & 34.4 & 21.5 & 14.6 & 10.5 & 13.8 & 27.9 & -- & -- &-- \\
        CMCL & 34.4 & 21.7 & 14.0 & 9.7 & 13.3 & 28.1 & -- & -- &-- \\
        PPKED & 36.0 & 22.4 & 14.9 & 10.6 & 14.9 & 28.4 & -- & -- &-- \\
        AlignTransformer & 37.8 & 23.5 & 15.6 & 11.2 & 15.8 & 28.3 & -- & -- &-- \\
        CA & 35.0 & 21.9 & 15.2 & 10.9  &15.1 & 28.3 & 35.2 & 29.8 & 30.3 \\
        LKBMA & 38.6 & 23.7 & 15.7 & 11.1 & --  & 27.4 & 42.0 & 33.9 & 35.2 \\
        M$^2$TR & 37.8 & 23.2 & 15.4 & 10.7 & 14.5 & 27.2 & 24.0 & 42.8 & 30.8 \\
        KnowMAT & 36.3 & 22.8 & 15.6 & 11.5 & -- & 28.4 & 45.8 & 34.8 & 37.1 \\
        RAMT & 36.2 & 22.9 & 15.7 & 11.3 & 15.3 & 28.4 & 38.0 & 34.2 & 33.5 \\
        CMM-RL & 38.1 & 23.2 & 15.5 & 10.9 & 15.1 & 28.7 & 34.2 & 29.4 & 29.2 \\
        CMCA & 36.0 & 22.7 & 15.6 & 11.7 & 14.8 & 28.7 & 44.4 & 29.7 & 35.6 \\
        KiUT & 39.3 & 24.3 & 15.9 & 11.3 & 16.0 & 28.5 & 37.1 & 31.8 & 32.1 \\
        DCL & -- & -- & -- & 10.9 & 15.0 & 28.4 & 47.1 & 35.2 & 37.3 \\
        MMTN & 37.9 & 23.8 & 15.9 & 11.6 & 16.1 & 28.3  & -- & -- &-- \\
        METrans & 25.0 & 16.9 & 12.4 & 15.2 & -- & 29.1 & 36.4 & 30.9 & 31.1 \\
        ORGAN & 38.6 & 25.6 & 17.2 & 12.3 & 16.2 & 29.3 & 41.6 & 41.8 & 38.5 \\
        COMG & 36.3 & 23.5 & 16.7 & 12.4 & 12.8 & 29.0 & -- & -- &-- \\
        MedM2G & 41.2 & 26.9 & 17.9 & 14.2 & -- & 30.9 & -- & -- &-- \\ 
        CvT2DistilGPT2 & 39.2 & 24.5 & 16.9 & 12.4 & 15.3 & 28.5 & 35.9 & 41.2 & 38.4 \\
        RGRG & 37.3 & 24.9 & 17.5 & 12.6 & 16.8 & 26.4  & 46.1 & \underline{47.5}  & \underline{44.7} \\
        BioViL-T & -- & -- & -- & 9.2 & -- & 29.6 & -- & -- & 17.5 \\
        VLCI & 40.0 & 24.5 & 16.5 & 11.9 & 15.0 & 28.0 & \underline{48.9} & 34.0 & 40.1 \\
        TiBiX & 32.4 & 23.4 & \underline{18.5} & \underline{15.7} & 16.2 & \underline{33.1} & 30.0 & 22.4 & 25.0\\
        RECAP & 42.9 & 26.7 & 17.7 & 12.5 & 16.8 & 28.8 & 38.9 & 44.3 & 39.3 \\
        MS-TF & \underline{43.6} & \underline{27.5} & 18.4 & 12.9 & \underline{17.7} & 30.5 & -- &-- & 41.1 \\
        \midrule
        Libra & \textbf{51.3} & \textbf{38.0} & \textbf{30.0} & \textbf{24.5} & \textbf{48.9} & \textbf{36.7} & \textbf{59.7} & \textbf{52.5} & \textbf{55.9} \\
        \bottomrule
    \end{tabular}
    \vspace{-5pt}
    \caption{\label{tab:appendix-nonllm}Findings Generation Performance of Libra and non-LLM-based Models. The best performances are highlighted in \textbf{bold}, and the second-best scores are \underline{underlined}. $^\text{‡}$ denotes results from  \citet{chen-etal-2021-cross-modal}, and `--' indicates missing data. These results are taken from the best performances reported in their original papers.}
    \vspace{-10pt}
\end{table}

To facilitate comparison with non-LLM-based models, we selected evaluation metrics commonly used in these studies. These include BLEU-$\{1,2,3,4\}$ \citep{10.3115/1073083.1073135}, METEOR (MTR) \citep{banerjee-lavie-2005-meteor}, and ROUGE-L (R-L) \citep{lin-2004-rouge}. For clinical metrics, we adopted the CheXbert \citep{irvin2019chexpertlargechestradiograph}, reporting Precision (P), Recall (R), and F$_1$.

\paragraph{Baseline} For performance evaluation, we compare our model with the following baselines: ST \citep{7298935}, ATT2IN \citep{8099614}, ADAATT \citep{8099828}, TopDown \citep{8578734}, R2Gen \citep{chen-etal-2020-generating}, R2GenCMN \citep{chen-etal-2021-cross-modal}, M$^2$TR \citep{nooralahzadeh-etal-2021-progressive-transformer}, CMCL \citep{liu-etal-2021-competence},PPKED \citep{9578840}, AlignTransformer \citep{10.1007/978-3-030-87199-4_7}, CA \citep{liu-etal-2021-contrastive}, LKBMA \citep{yang2022radiologyreportgenerationlearned}, KnowMAT \citep{Yang_2022}, XPRONET \citep{wang2022cross}, CMM-RL \citep{qin-song-2022-reinforced}, RAMT \citep{10.1109/TMM.2023.3273390}, CMCA \citep{song-etal-2022-cross}, KiUT \citep{huang2023kiutknowledgeinjectedutransformerradiology}, DCL \citep{li2023dynamicgraphenhancedcontrastive}, MMTN \citep{Cao_Cui_Zhang_Yu_Li_Xu_2023}, METrans \citep{wang2023metransformerradiologyreportgeneration}, ORGAN \citep{hou2023organobservationguidedradiologyreport}, COMG \citep{gu2023complexorganmaskguided}, BioViL-T \citep{bannur2023learningexploittemporalstructure}, RGRG \citep{Tanida_2023_CVPR}, RECAP \citep{hou-etal-2023-recap}, CvT2DistilGPT2 \citep{nicolson_improving_2023}, VLCI \citep{chen2024crossmodalcausalinterventionmedical}, TiBiX \citep{sanjeev2024tibixleveragingtemporalinformation}, MedM2G \citep{zhan2024medm2gunifyingmedicalmultimodal}, MS-TF \citep{10.1145/3664647.3681377}.

To ensure fairness, Libra also utilizes prior images, aligning with other models that leverage prior images or additional information. As demonstrated in Table~\ref{tab:appendix-nonllm}, Libra, similar to other LLM-based models, consistently outperforms non-LLM-based models. This advantage is largely attributed to advancements in LLMs and visual instruction tuning \citep{liu2023visualinstructiontuning}, enabling multimodal large language models (MLLMs) to achieve superior performance in RRG tasks.

\section{Additional Ablation Studies}\label{add-ablation}

\subsection{Impact of Temporal Information on Libra in RRG}

\begin{wraptable}{r}{0.48\linewidth}
    \small
    \centering
    \vfill
    \setlength{\tabcolsep}{2pt} 
    \begin{tabular}{l|c|c}
        \toprule
        \multirow{2}{*}{\textbf{Metric}} & \multicolumn{2}{c}{\textbf{Libra}}\\
        \cmidrule{2-3}
        & \textbf{w/o} prior & \textbf{w/} prior (\%) \\  
        \midrule
        \textbf{Lexical:} & & \\
        \hspace{1em} ROUGE-L & 36.17 &  36.66 (+1.35\%) \\
        \hspace{1em} BLEU-1 & 51.20 &  51.25 (+0.10\%) \\
        \hspace{1em} BLEU-4 & 24.33 &  24.54 (+0.86\%) \\
        \hspace{1em} METEOR & 48.69 &  48.90 (+0.43\%)\\ 
        \hspace{1em} BERTScore & 61.94 &  62.50 (+0.90\%) \\
        \midrule
        \rowcolor{lightgray}
        \hspace{1em} F1$_\text{temp}$ & 32.72 &  35.34 (+8.00\%) \\
        \midrule
         \textbf{Clinical:} & & \\
        \hspace{1em} RadGraph-F1 & 32.42 &  32.87(+1.39\%) \\
        \hspace{1em} RG$_{\text{ER}}$ & 36.92 &  37.57(+1.76\%) \\
        \hspace{1em} RadCliQ$_{\text{0}}$($\downarrow$)  & 2.76 &  2.72 (+1.45\%) \\
        \hspace{1em} CheXbert vector & 46.31  &  46.85 (+1.17\%) \\
        \hspace{1em} \textit{CheXpert-F1:} & & \\
        \hspace{2em} Micro-F1-14 & 55.25 &  55.87 (+1.12\%) \\
        \hspace{2em} Macro-F1-14 & 40.15 &  40.38(+0.57\%) \\
        \hspace{2em} Micro-F1-5  & 58.93 &  60.07(+1.93\%) \\
        \hspace{2em} Macro-F1-5  & 52.61 &  53.75(+2.17\%) \\ 
        \bottomrule
    \end{tabular}
    \vspace{-5pt}
    \caption{\label{tab:q1-original}Ablation results for Libra without \textbf{(w/o)} and with \textbf{(w/)} the prior image. Values in (\%) indicate the percentage improvement.}
\end{wraptable}

Temporal information is embedded in paired images and referenced in the corresponding radiology reports, capturing changes over time through references to prior symptoms and their progression, as discussed in Appx.~\ref{research-object-ti}. As shown in Table~\ref{tab:datasets}, \textbf{86\%} of the test data includes prior images, providing a solid foundation for evaluating the impact of temporal information.  

During training, Libra integrates the ability to perceive and utilise temporal information into its architecture. To evaluate whether Libra effectively leverage temporal information during inference, we assess its performance using prior images when available as references to determine their impact on the overall capability. 

In Table~\ref{tab:q1-original}, the inclusion of prior images substantially enhances Libra's performance across all metrics. Notably, clinical scores exhibit greater improvements compared to lexical scores, underscoring the importance of temporal information in generating high-quality medical reports beyond merely improving linguistic fluency.

The F1$_\text{temp}$ score shows the most substantial improvement, with an increase of $\textbf{8\%}$, highlighting Libra's capability to effectively leverage temporal changes provided by prior images. These results validate the role of temporal information in enhancing the quality of the generated \textit{Findings} section and improving Libra's overall performance in RRG tasks.

\subsection{Impact of the Temporal Alignment Connector under General-Domain Pre-trained Models}

\begin{table}[ht]

    \vspace{2pt}
    \small
    \centering
    \setlength{\tabcolsep}{8pt}
    \begin{tabular}{l|c|cccc}
        \toprule
        \textbf{Metric} & \textbf{\textit{Libra-b}} & \textbf{w/o TFM} & \textbf{w/o LFE} & \textbf{w/o PIPB} & \textbf{w/o TAC} \\  
        \midrule
        \textbf{Lexical:} & & & & \\
        \hspace{1em} ROUGE-L & 27.26 & 26.80 (-1.69\%) & 26.57 (-2.53\%) & 27.00 (-0.95\%) & 24.58 (-9.83\%) \\
        \hspace{1em} BLEU-1 & 34.94 & 33.61 (-3.81\%) & 33.68 (-3.61\%) & 34.47 (-1.35\%) & 31.40 (-10.13\%) \\
        \hspace{1em} BLEU-4 & 11.74 & 10.97 (-6.56\%) & 10.94 (-6.81\%) & 11.57 (-1.45\%) & 8.89 (-24.28\%) \\
        \hspace{1em} METEOR & 35.37 & 34.50 (-2.46\%) & 34.30 (-3.03\%) & 34.93 (-1.24\%) & 32.41 (-8.37\%) \\ 
        \hspace{1em} BERTScore & 55.51 & 54.97 (-0.97\%) & 54.75 (-1.37\%) & 55.26 (-0.45\%) & 53.07 (-4.40\%) \\
        \midrule
        \rowcolor{lightgray}
        \hspace{1em} F1$_\text{temp}$ & 24.77 & 23.54 (-4.97\%) & 24.00 (-3.11\%) & 24.68 (-0.36\%) & 22.52 (-9.08\%) \\
        \midrule
        \textbf{Clinical:} & & & & \\
        \hspace{1em} RadGraph-F1 & 21.67 & 21.06 (-2.81\%) & 20.73 (-4.34\%) & 21.35 (-1.48\%) & 19.77 (-8.77\%) \\
        \hspace{1em} RG$_{\text{ER}}$ & 26.28 & 25.45 (-3.16\%) & 25.14 (-4.34\%) & 25.84 (-1.67\%) & 23.74 (-9.67\%) \\
        \hspace{1em} RadCliQ$_{\text{0}}$ ($\downarrow$) & 3.17 & 3.20 (-0.95\%) & 3.22 (-1.58\%) & 3.18 (-0.32\%) & 3.27 (-3.15\%) \\
        \hspace{1em} CheXbert vector & 39.58 & 38.74 (-2.12\%) & 38.37 (-3.06\%) & 39.49 (-0.23\%) & 37.56 (-5.10\%) \\
        \hspace{1em} \textit{CheXpert-F1:} & & & & \\
        \hspace{2em} Micro-F1-14 & 49.06 & 47.68 (-2.81\%) & 47.57 (-3.04\%) & 48.40 (-1.35\%) & 46.57 (-5.08\%) \\
        \hspace{2em} Macro-F1-14 & 33.07 & 31.60 (-4.45\%) & 31.78 (-3.90\%) & 31.78 (-3.90\%) & 31.27 (-5.44\%) \\
        \hspace{2em} Micro-F1-5 & 54.55 & 52.14 (-4.42\%) & 52.94 (-2.95\%) & 53.10 (-2.66\%) & 50.72 (-7.02\%) \\
        \hspace{2em} Macro-F1-5 & 47.24 & 44.28 (-6.27\%) & 44.39 (-6.04\%) & 44.68 (-5.42\%) & 43.48 (-7.96\%) \\
        \bottomrule
    \end{tabular}
    \vspace{-5pt}
    \caption{ \label{tab:AC-1}Results of ablation experiments for the Temporal Alignment Connector. `$\downarrow$' indicates that lower is better. Values in (\%) indicate the percentage decrease compared with the \textit{Libra-b}.}
    \vspace{-15pt}
\end{table}

Domain-specific pre-trained models (i.e., RAD-DINO \citep{pérezgarcía2024raddinoexploringscalablemedical} and Meditron \citep{chen2023meditron70bscalingmedicalpretraining}) inherently incorporate domain-specific knowledge, such as phrasing conventions, pronoun usage, and even temporal information embedded in the training corpus. To isolate the structural impact of TAC, we used a general-domain image encoder (DINOv2 \citep{oquab2024dinov2learningrobustvisual}) and a LLM (Vicuna-7B-v1.5 \citep{vicuna2023}), allowing the structural enhancements of TAC to be observed more directly. 

We replicated the first ablation setup from Sec.~\ref{ablation-q1}. We first conducted a baseline experiment, referred to as \textit{Libra-b}, by fine-tuning only the adapter for the \textit{Findings} generation task. As shown in Table~\ref{tab:AC-1}, we then conducted ablation studies by sequentially removing different components from the model, including the Temporal Fusion Module (TFM), Layerwise Feature Extractor (LFE), Prior Image Prefix Bias (PIPB), and the entire TAC. Removing TFM restricts the model to processing only the current image, using a configuration similar to LLaVA \citep{liu2023visualinstructiontuning}, but with a four-layer MLP to align the image feature with the LLM's hidden dimensions. Notably, without TFM, the model cannot process prior images or dummy prior images, and is limited to only the current image as input. Without LFE, the model follows the LLaVA setup, using the penultimate layer of the image encoder to process single or paired images.

The ablation results are consistent with those observed using domain-specific models, as presented in Table~\ref{tab:q1}. Removing any TAC submodule led to declines across all metrics. Specifically, removing TFM caused a notable drop in the F1$_\text{temp}$ score ($\downarrow >$4\%), emphasising its role in capturing temporal information. The absence of LFE significantly reduced RadGraph-related scores, demonstrating its importance for detailed image feature extraction. PIPB removal primarily impacted clinical metrics, while removing the entire TAC resulted in substantial declines across all metrics. These findings reaffirm the critical role of TAC in integrating image details and temporal information effectively.

\subsection{Impact of the Temporal Alignment Connector After the Second-Stage Fine-tuning}\label{add-ablation-3}

\begin{table}[ht]
    \vspace{-10pt}
    \small
    \centering
    \setlength{\tabcolsep}{8pt}
    \begin{tabular}{l|c|cccc}
        \toprule
        \textbf{Metric} & \textbf{\textit{Libra-2}} & \textbf{w/o TFM} & \textbf{w/o LFE} & \textbf{w/o PIPB} & \textbf{w/o TAC} \\  
        \midrule
        \textbf{Lexical:} & & & & \\
        \hspace{1em} ROUGE-L & 35.31 & 35.16 (-0.42\%) & 35.09 (-0.64\%) & 35.23 (-0.23\%) & 34.41 (-2.55\%) \\
        \hspace{1em} BLEU-1 & 49.92 & 49.44 (-0.97\%) & 49.47 (-0.90\%) & 49.75 (-0.34\%) & 48.61 (-2.63\%) \\
        \hspace{1em} BLEU-4 & 23.05 & 22.67 (-1.66\%) & 22.65 (-1.75\%) & 22.97 (-0.35\%) & 21.51 (-6.70\%) \\
        \hspace{1em} METEOR & 47.99 & 47.69 (-0.62\%) & 47.62 (-0.77\%) & 47.84 (-0.31\%) & 46.95 (-2.16\%) \\ 
        \hspace{1em} BERTScore & 61.28 & 61.13 (-0.24\%) & 61.07 (-0.34\%) & 61.21 (-0.12\%) & 60.60 (-1.12\%) \\
        \midrule
        \rowcolor{lightgray}
        \hspace{1em} F1$_\text{temp}$ & 33.52 & 33.10 (-1.27\%) & 33.25 (-0.79\%) & 33.49 (-0.09\%) & 32.73 (-2.36\%) \\
        \midrule
        \textbf{Clinical:} & & & & \\
        \hspace{1em} RadGraph-F1 & 30.77 & 30.55 (-0.72\%) & 30.43 (-1.10\%) & 30.65 (-0.40\%) & 30.07 (-2.27\%) \\
        \hspace{1em} RG$_{\text{ER}}$ & 35.44 & 35.16 (-0.79\%) & 35.05 (-1.10\%) & 35.29 (-0.42\%) & 34.55 (-2.51\%) \\
        \hspace{1em} RadCliQ$_{\text{0}}$ ($\downarrow$) & 2.83 & 2.84 (-0.35\%) & 2.84 (-0.35\%) & 2.85 (-0.71\%) & 2.85 (-0.71\%) \\
        \hspace{1em} CheXbert vector & 45.32 & 45.08 (-0.53\%) & 44.97 (-0.77\%) & 45.27 (-0.11\%) & 44.73 (-1.30\%) \\
        \hspace{1em} \textit{CheXpert-F1:} & & & & \\
        \hspace{2em} Micro-F1-14 & 54.11 & 53.73 (-0.70\%) & 53.70 (-0.76\%) & 54.00 (-0.20\%) & 53.41 (-1.30\%) \\
        \hspace{2em} Macro-F1-14 & 37.16 & 36.74 (-1.13\%) & 36.78 (-1.02\%) & 36.79 (-1.00\%) & 36.64 (-1.40\%) \\
        \hspace{2em} Micro-F1-5 & 58.76 & 58.10 (-1.12\%) & 58.32 (-0.75\%) & 58.36 (-0.68\%) & 57.65 (-1.89\%) \\
        \hspace{2em} Macro-F1-5 & 51.99 & 51.16 (-1.60\%) & 51.19 (-1.54\%) & 51.27 (-1.38\%) & 50.87 (-2.15\%) \\
        \bottomrule
    \end{tabular}
    \vspace{-5pt}
    \caption{\label{tab:appenidx-c3}Results of ablation experiments for the Temporal Alignment Connector after the second stage. `$\downarrow$' indicates that lower is better. Values in (\%) indicate the percentage decrease compared with the \textit{Libra-2}.}
    \vspace{-10pt}
\end{table}

To further evaluate the impact of the Temporal Alignment Connector (TAC) on Libra's performance, we followed the setup of the first ablation study in Sec.~\ref{ablation-q2}. After the first stage of alignment, the model underwent a second stage of fine-tuning. This stage was designed to optimise the model's performance on the \textit{Findings} section generation task by leveraging the aligned visual and textual features learned during the initial stage.

In this phase, we applied Low-Rank Adaptation (LoRA) \citet{hu2021loralowrankadaptationlarge} to fine-tune the pre-trained LLM (Meditron \citet{chen2023meditron70bscalingmedicalpretraining}), while keeping the visual encoder (RAD-DINO \citet{pérezgarcía2024raddinoexploringscalablemedical}) and TAC weights frozen. The baseline for this experiment is \textit{Libra-2} (in Table~\ref{tab:appenidx-c3}), which is derived from \textit{Libra-1} (in Table~\ref{tab:q1}) after undergoing LoRA fine-tuning.

We conducted ablation studies by progressively removing different TAC components, including TFM, LFE, the Prior Image Prefix Bias (PIPB), and the entire TAC. Results consistently showed declines across all metrics compared to \textit{Libra-2}, mirroring the trends observed in Sec.~\ref{ablation-q1}. This reinforces that the performance improvements brought by TAC are stable and unaffected by changes in training stages. It further confirms that TAC has embedded the capability to process temporal information within the model.

\subsection{Robustness Evaluation of the Temporal Alignment Connector}

To evaluate the robustness of the Temporal Alignment Connector (TAC), we introduced an additional round of LoRA fine-tuning to induce overtraining. Following the setup in Appx.~\ref{add-ablation-3}, after integrating the first LoRA weights, a new set of LoRA adapters was reinitialised for the LLM and trained for one epoch under the same second-stage fine-tuning configuration. The baseline for this experiment is \textit{Libra-3} (as shown in Table~\ref{tab:appenidx-c4}), which is derived from \textit{Libra-2} (illustrated in Table~\ref{tab:appenidx-c3}) following this additional fine-tuning step.

\begin{table}[ht]
    \vspace{-5pt}
    \small
    \centering
    \setlength{\tabcolsep}{8pt}
    \begin{tabular}{l|c|cccc}
        \toprule
        \textbf{Metric} & \textbf{\textit{Libra-3}} & \textbf{w/o TFM} & \textbf{w/o LFE} & \textbf{w/o PIPB} & \textbf{w/o TAC} \\  
        \midrule
        \textbf{Lexical:} & & & & \\
        \hspace{1em} ROUGE-L & 35.58 & 35.53 (-0.14\%) & 35.51 (-0.21\%) & 35.55 (-0.08\%) & 35.28 (-0.86\%) \\
        \hspace{1em} BLEU-1 & 49.54 & 49.38 (-0.32\%) & 49.39 (-0.30\%) & 49.48 (-0.11\%) & 49.10 (-0.88\%) \\
        \hspace{1em} BLEU-4 & 23.61 & 23.48 (-0.55\%) & 23.47 (-0.58\%) & 23.58 (-0.12\%) & 23.07 (-2.28\%) \\
        \hspace{1em} METEOR & 47.61 & 47.51 (-0.21\%) & 47.49 (-0.26\%) & 47.56 (-0.10\%) & 47.26 (-0.73\%) \\ 
        \hspace{1em} BERTScore & 61.54 & 61.49 (-0.08\%) & 61.47 (-0.11\%) & 61.52 (-0.04\%) & 61.31 (-0.37\%) \\
        \midrule
        \rowcolor{lightgray}
        \hspace{1em} F1$_\text{temp}$ & 33.51 & 33.37 (-0.42\%) & 33.42 (-0.27\%) & 33.50 (-0.03\%) & 33.24 (-0.79\%) \\
        \midrule
        \textbf{Clinical:} & & & & \\
        \hspace{1em} RadGraph-F1 & 29.82 & 29.75 (-0.24\%) & 29.71 (-0.37\%) & 29.78 (-0.13\%) & 29.59 (-0.76\%) \\
        \hspace{1em} RG$_{\text{ER}}$ & 35.60 & 35.51 (-0.26\%) & 35.47 (-0.37\%) & 35.55 (-0.14\%) & 35.30 (-0.84\%) \\
        \hspace{1em} RadCliQ$_{\text{0}}$ ($\downarrow$) & 2.91 & 2.92 (-0.34\%) & 2.92 (-0.34\%) & 2.91 ( -- ) & 2.93 (-0.68\%) \\
        \hspace{1em} CheXbert vector & 44.77 & 44.69 (-0.18\%) & 44.65 (-0.26\%) & 44.75 (-0.04\%) & 44.57 (-0.45\%) \\
        \hspace{1em} \textit{CheXpert-F1:} & & & & \\
        \hspace{2em} Micro-F1-14 & 52.45 & 52.33 (-0.23\%) & 52.32 (-0.25\%) & 52.41 (-0.08\%) & 52.22 (-0.44\%) \\
        \hspace{2em} Macro-F1-14 & 30.77 & 30.65 (-0.38\%) & 30.66 (-0.34\%) & 30.67 (-0.33\%) & 30.63 (-0.47\%) \\
        \hspace{2em} Micro-F1-5 & 54.42 & 54.22 (-0.38\%) & 54.28 (-0.25\%) & 54.30 (-0.23\%) & 54.08 (-0.63\%) \\
        \hspace{2em} Macro-F1-5 & 44.58 & 44.34 (-0.54\%) & 44.35 (-0.52\%) & 44.37 (-0.46\%) & 44.26 (-0.72\%) \\
        \bottomrule
    \end{tabular}
    \vspace{-5pt}
    \caption{\label{tab:appenidx-c4}Results of ablation experiments for the Temporal Alignment Connector with additional LoRA fine-tuning after the second stage. `$\downarrow$' indicates that lower is better. Values in (\%) indicate the percentage decrease compared with the \textit{Libra-3}.}
    \vspace{-10pt}
\end{table}


The results reveal that, compared to \textit{Libra-2}, \textit{Libra-3} exhibits minimal changes in lexical scores, while clinical scores decline due to overfitting caused by the additional fine-tuning. Notably, the CheXpert \citep{smit-etal-2020-combining} (Macro-F1-[5/14]) scores exhibit the most influential reduction.

Despite this decline, ablation studies confirm that TAC's performance improvements remain robust, unaffected by variations in training strategies. This resilience stems from TAC’s ability to capture and retain temporal image representations during the initial training phase, which are preserved through subsequent fine-tuning.

These findings underscore TAC's reliability as a critical component for temporal information processing in RRG tasks. It ensures stability even under diverse training conditions.

\subsection{Impact of Radiology-Specific Pre-trained Models on Libra}

\begin{table}[ht]
    \vspace{-10pt}
    \small
    \centering
    \setlength{\tabcolsep}{8pt}
    \begin{tabular}{l|c|ccc}
        \toprule
        \textbf{Metric} & \textbf{Libra-1} & \textbf{w/o RadDINO} & \textbf{w/o Meditron} & \textbf{w/o RadDINO+Meditron} \\  
        \midrule
        \textbf{Lexical:} & & & & \\
        \hspace{1em} ROUGE-L & 27.56 & 27.66 (0.36\%) & 27.29 (-0.98\%) & 27.26 (-1.09\%) \\
        \hspace{1em} BLEU-1 & 34.84 & 35.32 (1.38\%) & 34.91 (0.20\%) & 34.94 (0.29\%) \\
        \hspace{1em} BLEU-4 & 11.51 & 12.56 (9.12\%) & 11.61 (0.87\%) & 11.74 (2.00\%) \\
        \hspace{1em} METEOR & 35.50 & 35.65 (0.42\%) & 35.53 (0.08\%) & 35.37 (-0.37\%) \\ 
        \hspace{1em} BERTScore & 55.87 & 55.89 (0.04\%) & 55.58 (-0.52\%) & 55.51 (-0.64\%) \\
        \midrule
        \rowcolor{lightgray}
        \hspace{1em} F1$_\text{temp}$ & 26.63 & 25.53 (-4.13\%) & 24.78 (-6.95\%) & 24.77 (-6.98\%) \\
        \midrule
        \textbf{Clinical:} & & & & \\
        \hspace{1em} RadGraph-F1 & 22.52 & 22.11 (-1.82\%) & 23.13 (2.71\%) & 21.67 (-3.77\%) \\
        \hspace{1em} RG$_{\text{ER}}$ & 27.32 & 26.72 (-2.20\%) & 27.53 (0.77\%) & 26.28 (-3.81\%) \\
        \hspace{1em} RadCliQ$_{\text{0}}$ ($\downarrow$) & 3.10 & 3.13 (-0.97\%) & 3.08 (0.65\%) & 3.17 (-2.26\%) \\
        \hspace{1em} CheXbert vector & 42.02 & 40.78 (-2.95\%) & 41.94 (-0.19\%) & 39.49 (-6.02\%) \\
        \hspace{1em} \textit{CheXpert-F1:} & & & & \\
        \hspace{2em} Micro-F1-14 & 52.84 & 51.55 (-2.44\%) & 51.45 (-2.63\%) & 49.06 (-7.15\%) \\
        \hspace{2em} Macro-F1-14 & 36.87 & 34.58 (-6.21\%) & 37.20 (0.90\%) & 33.07 (-10.31\%) \\
        \hspace{2em} Micro-F1-5 & 56.63 & 55.00 (-2.88\%) & 55.39 (-2.19\%) & 54.55 (-3.67\%) \\
        \hspace{2em} Macro-F1-5 & 49.33 & 47.26 (-4.20\%) & 47.62 (-3.47\%) & 47.24 (-4.24\%) \\
        \bottomrule
    \end{tabular}
    \vspace{-5pt}
    \caption{\label{tab:appenidx-h5}Ablation results for radiology-specific pre-trained models in Libra. `$\downarrow$' indicates that lower is better. Values in (\%) indicate the percentage improvement compared to \textit{Libra-c}.}
    \vspace{-10pt}
\end{table}

Aligning radiology images with textual information is a key challenge in RRG tasks. To demonstrate the benefits of using radiology-specific pre-trained models for more accurate feature representation and improved MLLM performance, we initialised a Libra model with RadDINO, the TAC, and Meditron-7b, conducting the first stage of training, denoted as \textit{Libra-1} (This is consistent with the baseline setup of the previous ablation study in Sec.~\ref{ablation-q1}). Then we replaced the image encoder and LLM with their general-domain counterparts, DINOv2 and Vicuna-7B-v1.5, respectively. Finally, we replaced both components, which is also referred to as \textit{Libra-b} (in Table~\ref{tab:AC-1}). 

As shown in Table~\ref{tab:appenidx-h5}, substituting radiology-specific pre-trained models with general-domain models resulted in a notable decline in clinical scores, while the impact on lexical scores was minimal. Notably, replacing the radiology-specific image encoder caused a more pronounced decline in clinical metrics compared to replacing the language model. This suggests that accurate medical image representation provides greater benefits in RRG tasks, indicating the importance of incorporating domain-specific knowledge into pre-trained models to enhance Libra's performance.

\subsection{Incremental Component Analysis}



\begin{table}[ht]
    \begin{center}
    \setlength{\tabcolsep}{6pt} 
    \small
    \begin{tabular}{l|ccccccc|c}
        \toprule
        \multirow{1}{*}{\textbf{Metric}} & \multicolumn{7}{c|}{\textbf{Stage 1:} Temporal Feature Alignment}& \multicolumn{1}{c}{\textbf{Stage 2}}\\
        \cmidrule(lr){2-8}\cmidrule(lr){9-9}
        & \textbf{$^*$Initial} & \textbf{$^\text{/}$DINO} & \textbf{$^\text{+}$LFE} &  \textbf{$^\text{+}$TFM} &  \textbf{$^\text{/}$RAD-DINO} & \textbf{$^\text{/}$Meditron} &  \textbf{$^\text{‡}$Dataset} &  \textbf{Libra} \\  
        \midrule
        \textbf{Lexical:} & & & & & & &\\
        \hspace{1em} ROUGE-L & 23.77 & 24.58  & 26.57  & 27.26 & 27.29 & \underline{27.56} & 27.27 & \textbf{36.66}\\
        \hspace{1em} BLEU-1 & 31.48 & 31.40  & 33.68  & 34.94 & 34.91 & 34.84 & \underline{41.24} & \textbf{51.25}  \\
        \hspace{1em} BLEU-4 & 8.41 & 8.89 & 10.94  & 11.74 & 11.61 & 11.51 & \underline{13.59} & \textbf{24.54} \\
        \hspace{1em} METEOR & 32.1 & 32.41 & 34.3  & 35.37 & 35.53 &35.50 & \underline{39.44} & \textbf{48.90}\\ 
        \hspace{1em} BERTScore & 52.76 & 53.07 & 54.75  & 55.51 & 55.58 & 55.87 & \underline{56.00} & \textbf{62.50} \\
        \midrule
        \rowcolor{lightgray}
        \hspace{1em} F1$_\text{temp}$ & 21.60 & 22.52 & 24.00  & 24.77 & 24.78 & \underline{26.63} & 24.80 & \textbf{35.34}\\
        \midrule
        \textbf{Clinical:} & & & & & & &  \\
        \hspace{1em} RadGraph-F1 & 18.58 & 19.70 & 20.73  & 21.67 & \underline{23.13} & 22.52 & 20.45 & \textbf{32.87} \\
        \hspace{1em} RG$_{\text{ER}}$ & 23.05 & 23.74 & 25.14  & 26.28 & \underline{27.53} & 27.32 & 25.19 & \textbf{37.57} \\
        \hspace{1em} RadCliQ$_{\text{0}}$($\downarrow$)& 3.35 & 3.26 & 3.22 & 3.17 & \underline{3.08} & 3.10 & 3.31 & \textbf{2.72}  \\
        \hspace{1em} CheXbert vector & 35.59 & 37.94 & 38.37  & 39.49 & 41.94 & \underline{42.02} & 35.33 & \textbf{46.85}  \\
        \hspace{1em} \textit{CheXpert-F1:} & & & & & && \\
        \hspace{2em} Micro-F1-14 & 44.75 & 46.57 & 47.57  & 49.06 & 51.45 & \underline{52.48} & 43.63 & \textbf{55.87}\\
        \hspace{2em} Macro-F1-14 & 25.13 & 31.27  & 31.07  & 33.07 & \underline{37.20} & 36.87 & 25.68 & \textbf{40.38} \\
        \hspace{2em} Micro-F1-5 & 45.97 & 50.72  & 52.94  & 54.55 & 55.39 & \underline{56.63} & 49.75 & \textbf{60.07}\\
        \hspace{2em} Macro-F1-5 & 36.55 & 43.48  & 44.39 & 47.24 & 47.62 & \underline{49.33} & 40.40 & \textbf{53.75}\\  
        \bottomrule
    \end{tabular}
    \vspace{-5pt}
    \caption{\label{tab:appendixablation}Results of ablation experiments for key components of Libra on \textit{Findings} section generation performance. $^*$ indicates our initialised model. $^\text{/}$ denotes component replacement. $^\text{+}$ signifies structural addition. $^\text{‡}$ represents dataset configuration. The best performances are highlighted in \textbf{bold}, and the second-best scores are \underline{underlined}. `$\downarrow$' indicates that lower is better.}
    \end{center}
    \vspace{-15pt}
\end{table}

We conducted an incremental study to evaluate the effectiveness of each component in Libra's architecture. Starting with a baseline model similar to LLaVA—comprising a pre-trained CLIP image encoder, a randomly initialised four-layer MLP adapter, and Vicuna-7B-v1.5 as the LLM—we trained the adapter on the \textit{Findings} section generation task.

Improvements were introduced incrementally, as summarised in Table~\ref{tab:appendixablation}. First, we replaced the image encoder with DINOv2. Next, we incorporated the LFE (prefix module of TAC) and subsequently added the TFM (suffix module), completing the TAC connector. We then replaced the image encoder and LLM with RAD-DINO and Meditron, respectively. The dataset for the first stage was expanded, and final fine-tuning was conducted for downstream tasks to produce Libra.

With each enhancement, the model's performance improved, demonstrating the critical role of each component. Notably, the addition of the TFM during the alignment stage provided the most significant improvement, showcasing its ability to capture temporal information, which is essential for the RRG task.

However, data expansion in the first stage led to improved lexical scores but a slight decline in clinical metrics, likely due to the VQA task's focus on fine-grained grounded information rather than holistic report generation, as mentioned in Sec.~\ref{ablation-q2}. This shift also affected the F1$_\text{temp}$ score, as temporal entities are often linked to specific symptoms. These declines were subsequently addressed through second-stage fine-tuning, resulting in overall improved performance.

\begin{wraptable}{r}{0.5\textwidth}
    \vspace{-10pt}
    \centering
    \begin{minipage}[t]{0.5\textwidth} 
        \centering
        \vspace{8pt}
        \small
        \begin{tikzpicture}
            \begin{axis}[
                ylabel={F1$_\text{temp}$ score}, 
                symbolic x coords={$^*$Initial, $^\text{/}$DINO, $^\text{+}$LFE, $^\text{+}$TFM, $^\text{/}$RAD-DINO, $^\text{/}$Meditron, $^\text{‡}$Dataset, Libra}, 
                xtick=data, 
                ymin=20.0, ymax=38.0, 
                xticklabel style={font=\scriptsize, rotate=45, anchor=north east}, 
                xtick pos=lower, 
                ytick pos=left, 
                ymajorgrids=true, 
                grid style=dashed, 
                width=7.5cm, 
                height=7cm, 
                legend style={
                    font=\footnotesize,
                    at={(0.3,0.8)}, 
                    anchor = south, 
                    fill=white, 
                    draw=black 
                }
            ]
            \addplot[
                color=orange, 
                mark=triangle*,
                mark size=1pt,
                line width=1.3pt 
                ]
                coordinates {
                ($^*$Initial,20.73) ($^\text{/}$DINO,22.26) ($^\text{+}$LFE,23.97) ($^\text{+}$TFM,24.92) ($^\text{/}$RAD-DINO, 26.51) ($^\text{/}$Meditron,27.55) ($^\text{‡}$Dataset,24.92) (Libra, 36.33)
            };
            \addlegendentry{w/ prior}
            \addplot[
                color=blue, 
                mark=*,
                mark size=1pt,
                line width=1.3pt 
                ]
                coordinates {
                ($^*$Initial, 21.74) ($^\text{/}$DINO,24.17) ($^\text{+}$LFE,24.18) ($^\text{+}$TFM,23.83) ($^\text{/}$RAD-DINO,24.38) ($^\text{/}$Meditron,26.48) ($^\text{‡}$Dataset,24.07) (Libra,35.18)
            };
            \addlegendentry{w/o prior}
            \end{axis}
        \end{tikzpicture}
        \vspace{-10pt}
        \captionof{table}{\label{tab:apptemporal-f1}Results of ablation experiments for Libra on the F1$_\text{temp}$ score. Of the 2,461 official test samples, 2,117 include a prior image as a reference and 344 do not.}
        \vspace{-50pt}
    \end{minipage}
    
\end{wraptable}

\paragraph{Evaluation of Libra's Temporal Awareness} 

Another approach to investigating the model's ability to capture temporal information is to evaluate it separately within the test split based on the presence or absence of prior images, as shown in Table~\ref{tab:apptemporal-f1}. 

With the addition of the TFM, the model exhibited temporal awareness. It is worth noting that, for the first time, the F1$_\text{temp}$ score of samples with prior images surpassed those without, and this trend persisted through subsequent optimisations. This indicates that the structural enhancements have resulted in a sustained improvement in the model's temporal perception capabilities. An effective example is in Sec.~\ref{TA case}.
\vspace{50pt}

\section{Heatmap Analysis and Temporal Feature Representation}\label{heatmap-analysis}


\begin{figure}[ht]
    \centering
    \includegraphics[width=1.0\linewidth]{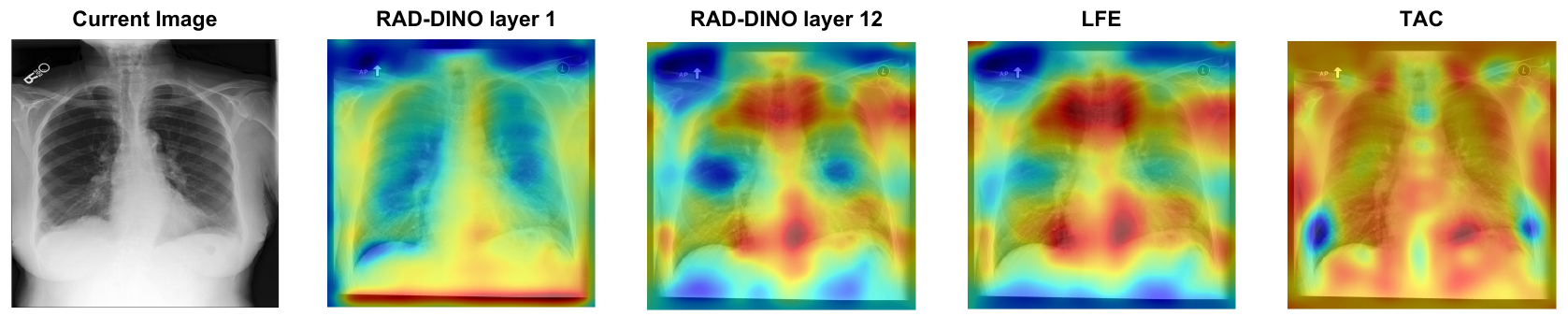}
    \vspace{-18pt}
    \caption{Heat map visualisation of image representations from different image encoder layers and the Temporal Alignment Connector (TAC), up-sampled using a Gaussian filter. Warm colours (red, yellow) indicate regions with higher weight allocations in the intermediate outputs of the ``hidden-state'' within the model blocks, while cool colours (blue, green) represent regions with lower weight.}
    \label{fig:heatmap-1}
    \vspace{-10pt}
\end{figure}

The heatmap in Figure~\ref{fig:heatmap-1} corresponds to the example in \textbf{\textit{(a)}} of Figure~\ref{fig:examples-section5}, where no prior image was used as a reference. It illustrates the clear differences in feature representations across layers of the RAD-DINO \citep{pérezgarcía2024raddinoexploringscalablemedical} image encoder. The shallow layers primarily capture the overall lung structure, while the deeper layers focus on specific disease regions. 

After passing through the Layerwise Feature Extractor (LFE), the image feature representations assign higher weights to larger symptom regions, achieving finer granularity. Following the Temporal Alignment Connector (TAC), the model integrates the weighted dummy prior image, producing a uniform feature distribution that reflects temporal information. This indicates no significant changes compared to the prior study and facilitates smoother image feature representations for downstream text generation by the LLM.

\begin{figure*}[ht]
    \centering
    \scriptsize
    \includegraphics[width=1.0\linewidth]{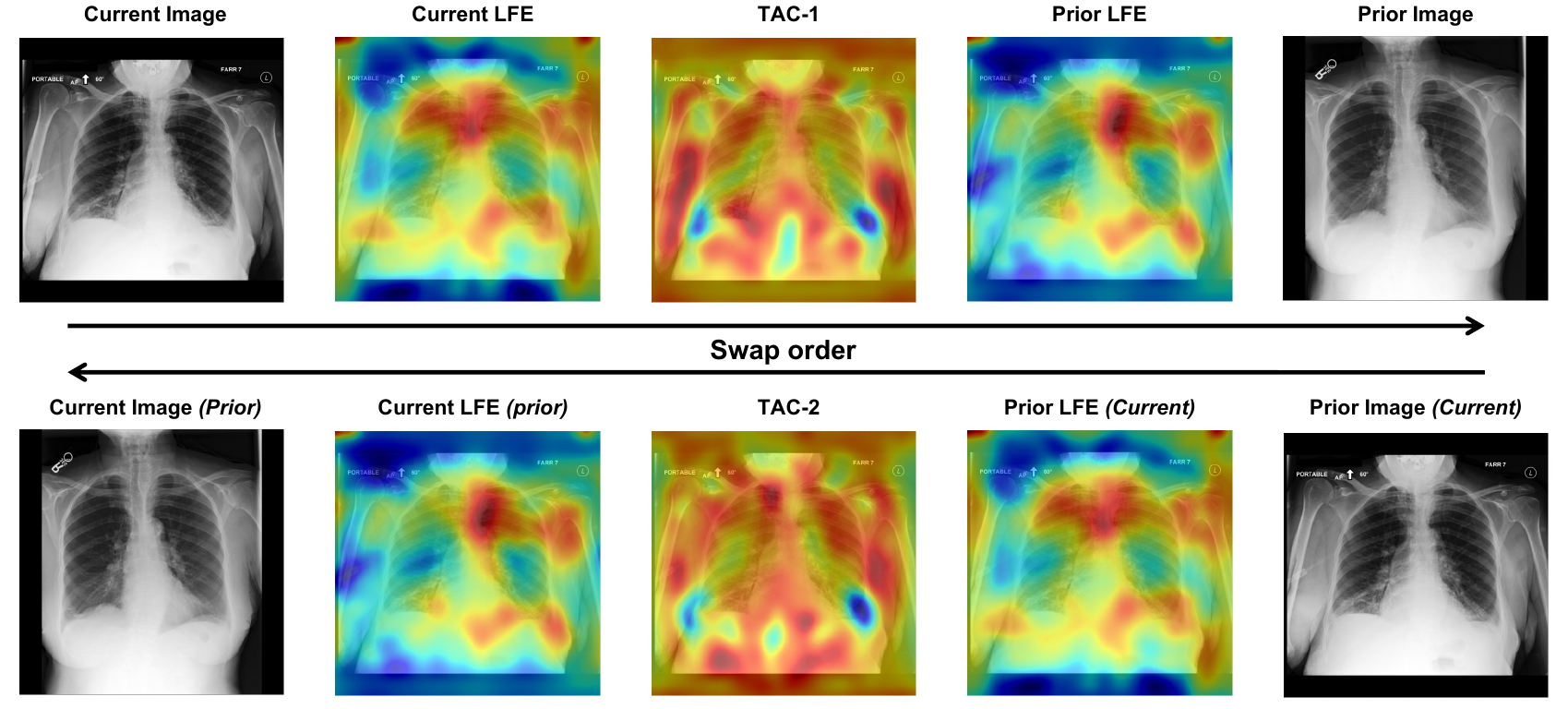}
    \vspace{-18pt}
    \caption{Heat map visualisation of image representations from the Temporal Alignment Connector (TAC), up-sampled using a Gaussian filter. The arrows (`$\rightarrow$') represent the direction of temporal information, pointing from the prior image to the true current image. Warm colours (red, yellow) indicate regions with higher weight allocations in the intermediate outputs of the ``hidden-state'' within the model blocks, while cool colours (blue, green) represent regions with lower weight.}
    \label{fig:heatmap-2}
    \vspace{-5pt}
\end{figure*}

The heatmap in Figure~\ref{fig:heatmap-2} corresponds to the example in \textbf{\textit{(b)}} of Figure~\ref{fig:examples-section5}, where a prior image is provided. After processing through the LFE, the model captures fine-grained feature representations in symptom areas. When processed by the TAC, these features are integrated with the differences between the two images, effectively reflecting temporal information, as demonstrated in TAC-1 (top) of Figure~\ref{fig:heatmap-2}.

When the image order is swapped, treating the prior image as the current image, the LFE output remains unchanged. However, comparing TAC-2 (bottom of Figure~\ref{fig:heatmap-2}) and TAC-1 outputs reveals significant differences in lung feature representations. This highlights the model's directional temporal perception and confirms that the TAC module effectively encodes temporal information from different time points, while the LFE focuses solely on image features without temporal encoding.

This behaviour aligns with the design of the TAC, where residual connections prioritise the current image as the main modality and the prior image as the auxiliary. Swapping the image order changes the main modality, altering the temporal state of symptoms in the generated report, such as reversing descriptions from ``improving'' to ``worsening,'' as discussed in Sec.~\ref{TA case}.

\section{Extended Discussion on Limitations}\label{limitations}
While our work represents a step forward in leveraging temporal information for radiology report generation,  it also has several limitations that warrant further exploration.

\paragraph{Handling Multiple Prior Scans} 
Our current model is designed to process a single prior scan alongside the current scan. While this approach aligns with standard clinical workflows, which typically prioritise the most recent prior study for comparisons, it overlooks scenarios where multiple prior scans could offer a richer temporal perspective. For instance, analysing a sequence of images spanning an extended period could provide deeper insights into gradual disease progression. Future efforts should focus on extending our framework to incorporate multiple prior scans efficiently, enabling a more nuanced understanding of temporal patterns in clinical data.

\paragraph{Temporal Information Beyond Image Comparisons} 
Currently, our model captures temporal information through paired image comparisons and corresponding textual reports. However, clinical assessments often draw upon a broader context, including historical notes, laboratory results, and other longitudinal patient data. Expanding our approach to integrate these diverse temporal data sources could facilitate a more holistic understanding of disease trajectories and patient history, significantly enhancing clinical applicability.

\paragraph{Sparse Temporal Data Challenges} 
In cases where prior scans are unavailable or minimally informative (e.g., taken within a short interval), our ``dummy prior image'' provides a workaround. However, the model's ability to interpret and generate meaningful outputs under these constraints may still be limited. Future research could focus on synthesising or imputing temporal context to enhance performance under these constraints.

\paragraph{Computational Complexity} 
The use of temporal alignment mechanisms and multi-layer feature integration increases computational demands, posing challenges for deployment in resource-constrained environments. Future optimisation efforts should focus on reducing computational overhead while maintaining performance.

\paragraph{Generalisability Across Modalities and Datasets} 
Our study is limited to frontal-view chest X-rays and the MIMIC-CXR dataset \citep{johnson2019mimic}. The applicability of our approach to other imaging modalities (e.g., CT, MRI) and datasets (e.g., CheXpert \citep{irvin2019chexpertlargechestradiograph}, PadChest \citep{bustos2020padchest}) remains unexplored. Future studies should assess the model's generalisability to a broader range of datasets and imaging contexts.
\vspace{5pt}
\\
Based on the identified limitations, we outline the following directions:

\begin{itemize}[align=right,itemindent=2em,labelsep=2pt,labelwidth=1em,leftmargin=0pt,nosep,topsep=2pt]
    \setlength{\itemsep}{2pt}
    \item Develop frameworks for integrating multiple prior scans with dynamic temporal reasoning to better capture longitudinal changes.
    \item Expand the model to incorporate multi-modal imaging and textual data for more comprehensive diagnostic insights.
    \item Investigate the integration of diverse temporal data sources, such as electronic health records (EHRs), to enhance clinical applicability.
    \item Exploring lightweight model architectures for faster inference while maintaining high performance.
\end{itemize}

These advancements aim to address the current limitations while broadening the applicability of temporal-aware multimodal models in radiology and other clinical domains.

\end{document}